\documentclass[conference]{IEEEtran}
\IEEEoverridecommandlockouts
\usepackage{cite}
\usepackage{amsmath,amssymb,amsfonts}
\usepackage{graphicx}
\usepackage{textcomp}
\usepackage{xcolor}
\def\BibTeX{{\rm B\kern-.05em{\sc i\kern-.025em b}\kern-.08em
    T\kern-.1667em\lower.7ex\hbox{E}\kern-.125emX}}

\usepackage{multirow}
\usepackage{graphicx} 
\usepackage{array}    
\usepackage{colortbl} 
\usepackage{makecell}
\usepackage{booktabs}
\usepackage{svg}
\usepackage{subcaption}
\usepackage{tabularx}
\usepackage{algorithm}
\usepackage{algpseudocode}
\usepackage{xcolor}
\usepackage{balance}

\begin{document}

\title{Evolving from Lessons: Skill-Augmented Table Graph Reasoning for Operation-wise Table Question Answering}

\author{\IEEEauthorblockN{Guixin Su, Qiankun Pi, Mayi Xu, Wenli Li, Ming Zhong, Yuanyuan Zhu, Jiawei Jiang, Tieyun Qian\textsuperscript{*} \thanks{*corresponding author.}}
\IEEEauthorblockA{\textit{School of Computer Science}, \textit{Wuhan University}, \textit{Wuhan, China}
}}

\maketitle

\begin{abstract}
Table Question Answering (TableQA) aims to reason over tables to answer user queries. Existing research treats all questions uniformly and evaluates solely through overall accuracy, obscuring a critical reality that LLMs excel at simple lookups yet struggle with complex operations like aggregation and arithmetic. To reveal this disparity, we introduce a novel \emph{Operation-wise TableQA} task with a fine-grained question taxonomy and release two datasets named WikiTQ-ow and TabFact-ow for evaluation. As for modeling bottlenecks, existing methods flatten tables into linearized texts, disrupting inherent structures and inducing the ``lost-in-the-middle'' issue, which poses a primary barrier to complex cross-row reasoning. Moreover, they typically reason from scratch, neglecting reusable patterns shared across similar operations. To address these limitations, we propose a Skill-augmented Table Graph Reasoning (SkillTGR) framework for self-evolving structured reasoning. Specifically, SkillTGR represents tables as attributed graphs with explicit row-column-cell structures, where LLMs plan and execute dynamic chains to retrieve evidence subgraphs for graph traversal reasoning. Based on this, SkillTGR builds a hierarchical SkillBank to distill reason trajectories into abstract skills under cognitive heuristics, then hybrid retrieves both successful and failed skills for contrastive augmented table graph reasoning, thereby enabling the continual self-evolution. Extensive experiments demonstrate that SkillTGR achieves superior performance with an average of 5.91\% overall and 5.06\% operation-wise improvement, also reducing 19.76\% token consumption and 27.64\% inference latency. Our codes and data will be released upon publication.
\end{abstract}


\section{Introduction}
As the fundamental task in table reasoning, Table Question Answering (TableQA) focuses on resolving natural language queries through evidence retrieval and reasoning over given tables, requiring models to understand tabular structures, precisely locate relevant entries, and perform multi-step numerical computations, which is widely applied in data engineering~\cite{AID-SQL-ICDE2025, Chat2DB-ICDE2025, FeVisQA-ICDE2025}, financial auditing~\cite{FinQA-EMNLP2021} and medical records analysis~\cite{EMRQA-EMNLP2018}.

Early studies mainly involve training or fine-tuning specialized TableQA models~\cite{BERT-NAACL2019, BART-ACL2020, T5-JMLR2020}. TAPAS~\cite{TAPAS-ACL2020} introduces weakly supervised pre-training to learn text-table correlations without logical forms, while TAPEX~\cite{TAPEX-ICLR2022} mitigates data scarcity by mimicking a neural SQL executor. TACUBE~\cite{TaCube-EMNLP2022} further enhances numerical reasoning through pre-computing question-aware data cubes. However, these methods exhibit suboptimal results when handling complex questions due to the limited reasoning capabilities of small-scale pre-trained models.

\begin{figure}
    \centering
    \includegraphics[width=1.0\linewidth]{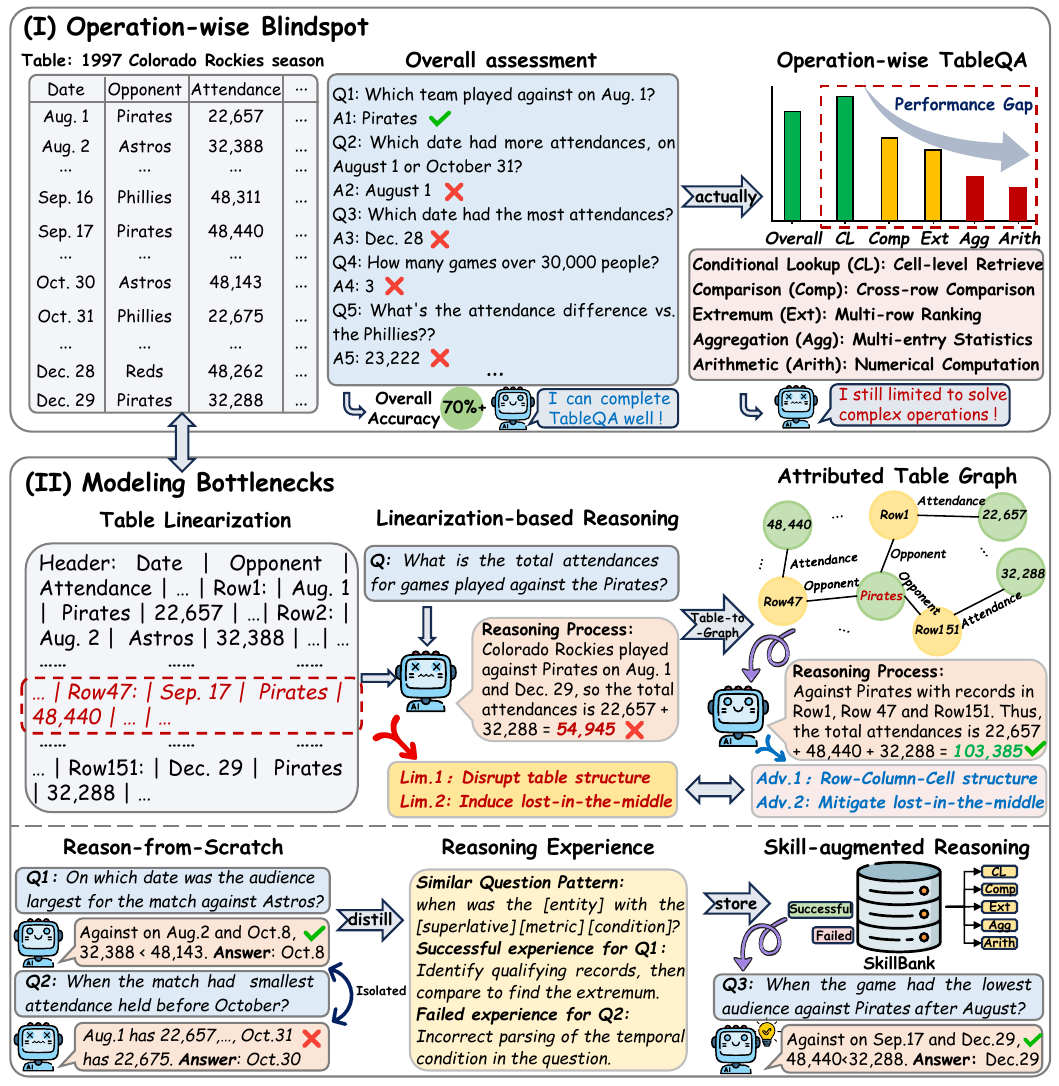}
    \caption{Operation-wise blindspot and modeling bottlenecks in existing research, and promising solution for TableQA.}
    \label{fig:introcuction}
\end{figure}

In recent years, large language models (LLMs) have demonstrated their remarkable capabilities in semantic comprehension and knowledge reasoning across diverse tasks~\cite{DatalogReasoning-ICDE2025, ChainsFormer-ICDE2025, MathReasoning-AAAI2026}. Consequently, recent research has resorted to leveraging LLMs to address the complex reasoning bottlenecks in TableQA.

An intuitive solution is to employ decomposition strategies that iteratively break down tables or questions into intermediate reasoning steps for progressive problem solving. For example, Binder~\cite{Binder-ICLR2023} utilizes LLMs to decompose complex questions into executable sub-programs (e.g., SQL or Python), and iteratively refine them before execution. Based on this, Dater~\cite{Dater-SIGIR2023} employs a parsing-execution-filling strategy for question decomposition and further leverages LLMs as versatile decomposers for sub-table retrieval. In addition, Chain-of-Table~\cite{Chain-of-Table-ICLR2024} dynamically evolves tables step-by-step through the reasoning chain to facilitate context-aware reasoning. However, decomposition-based methods often suffer from error propagation and insufficient interaction across reasoning steps. Therefore, several studies introduce multi-agent frameworks to collaboratively optimize the reasoning process. For instance, ReAcTable~\cite{ReAcTable-VLDB2024} follows the reason-then-action~\cite{ReAct-ICLR2023} paradigm to iteratively reason and execute external code tools, while Table-Critic~\cite{TableCritic-ACL2025} assigns LLMs as judge, critic, and refiner agents to progressively improve reasoning until convergence to correct solutions. Besides, some efforts~\cite{TrustUQA-AAAI2025, TIDE-ICLR2025} exploit the structured semantics of tables or questions to enhance evidence retrieval and reasoning reliability.

Despite the notable progress, existing studies still exhibit substantial limitations in terms of evaluation granularity and modeling paradigms, which naturally raise two key research questions as discussed below.

\emph{\textbf{RQ1: How does coarse-grained overall assessment obscure the operation-wise reasoning disparity of LLMs?}} Existing studies treat all questions uniformly and evaluate TableQA performance solely through the coarse-grained overall accuracy, obscuring a critical reality that LLMs exhibit markedly different capabilities across diverse operations. As illustrated in Fig. \ref{fig:introcuction} (I), previous methods tend to excel at simple questions involving conditional lookup, which requires the model to locate relevant entries through cell-level retrieval. In contrast, they still struggle with complex questions demanding cross-row reasoning across multiple entries, such as comparison, extremum and aggregation, as well as arithmetic questions requiring multi-step numerical computation. Therefore, coarse-grained evaluation obscures the operation-wise reasoning disparity of LLMs and fails to precisely identify reasoning bottlenecks for model improvement, highlighting the necessity of fine-grained evaluation with an operation-wise question taxonomy for comprehensive assessment.

\emph{\textbf{RQ2: What modeling bottlenecks underlie the operation-wise reasoning disparity of existing methods?}} As depicted in Fig. \ref{fig:introcuction} (II), existing methods suffer from inherent deficiencies in both table representation and reasoning paradigm. In terms of table representation, these methods typically linearize tables into flat sequences as long-context inputs, inevitably disrupting the row-column-cell structural semantics and impairing cross-row relational reasoning. In addition, such linearization further induces the ``lost-in-the-middle'' issue, where LLMs over-attend to boundary tokens while neglecting critical intermediate evidence, thereby hindering multi-row information retrieval for complex operations. On the reasoning side, prevailing methods uniformly follow the reason-from-scratch paradigm to infer each question independently without leveraging any historical experience. Thus, reasoning patterns for structurally similar operation-wise questions can neither be effectively reused nor transferred across instances, imposing constraints on both reasoning efficiency and answer accuracy.

For the \textbf{\emph{RQ1}}, we introduce a novel \emph{Operation-wise TableQA} task to investigate how LLMs perform across a fine-grained question taxonomy with five core operations rather than limited overall assessment. Specifically, \emph{Conditional Lookup} retrieves target entries via cell-level matching, while \emph{Comparison} and \emph{Extremum} require cross-row relational reasoning to identify differences or boundary values. \emph{Aggregation} involves jointly processing multiple entries for statistical summarization, whereas \emph{Arithmetic} further demands multi-step numerical computation over retrieved evidence. To support systematic evaluation, we construct two refined datasets named WikiTQ-ow and TabFact-ow, by annotating questions with operation types for fine-grained operation-wise analysis.

For the \textbf{\emph{RQ2}}, we propose a Skill-augmented Table Graph Reasoning (SkillTGR) framework for self-evolving structured reasoning. SkillTGR transforms tables into attributed graphs with explicit row-column-cell structures, enabling LLMs to employ predefined operators to plan and execute dynamic chains for evidence subgraph retrieval and graph traversal reasoning, thereby mitigating both structural semantic loss and the ``lost-in-the-middle'' issue inherent in table linearization. Furthermore, SkillTGR constructs a hierarchical SkillBank to distill historical trajectories into reusable skills under cognitive heuristics, and performs hybrid retrieval of both successful and failed skills for contrastive augmented reasoning, effectively breaking the reason-from-scratch limitation and supporting continuous self-evolution across diverse instances.

Our contributions can be concluded as follows:
\begin{itemize}
    \item To the best of our knowledge, we are the first to investigate the TableQA task in the fine-grained perspective with an operation-wise question taxonomy rather than limited overall assessment, which explores how LLMs perform in five core types of table reasoning. Accordingly, we release two refined benchmark datasets named WikiTQ-ow and TabFact-ow for evaluation, also reproduce and construct the baseline system with diverse modeling paradigms for operation-wise comparison.

    \item We propose a novel Skill-augmented Table Graph Reasoning (SkillTGR) framework for self-evolving structured reasoning. SkillTGR represents tables as attributed graphs, enabling structure-aware evidence retrieval and graph traversal reasoning. Moreover, SkillTGR constructs a hierarchical SkillBank to distill historical trajectories into reusable skills for subsequent reasoning, thereby progressively evolving reasoning capability across diverse operation-wise questions.  
    
    \item Extensive experiments on WikiTQ-ow and TabFact-ow across two LLMs demonstrate that SkillTGR consistently achieves the state-of-the-art performance, yielding average improvements of 5.91\% in overall accuracy and 5.06\% in operation-wise accuracy over competitive baselines, while reducing token consumption by 19.76\% and inference latency by 27.64\%, validating both the effectiveness and efficiency of our proposed framework.
\end{itemize}

\section{Related Work}
This section reviews related work from two relevant directions: \emph{Table Question Answering} and \emph{Self-Evolution for Large Language Models}. Representative studies from both areas are briefly discussed below.

\subsection{Table Question Answering}
As the core task in table reasoning, Table Question Answering requires models to jointly understand table structures and contents while parsing complex question intents, has evolved through several paradigms.

Early work primarily focused on developing specialized models through fine-tuning pre-trained language models~\cite{TaBERT-ACL2020, TAPEX-ICLR2022, PASTA-EMNLP2022}. Benefiting from the remarkable instruction-following capabilities of large language models (LLMs)~\cite{Fewshot-NeurIPS2020, Zero-Shot-NeurIPS2022, Self-Consistency-ICLR2023}, recent studies have shifted toward in-context learning paradigms~\cite{TableRAG-NeurIPS2024, TaPERA-ACL2024, RoT-EMNLP2025, TAREX-AAAI2026}. Inspired by step-wise decomposition strategies for progressive reasoning~\cite{Least-to-Most-ICLR2023, DecomP-ICLR2023, ST-Raptor-SIGMOD2025, SIRV-AAAI2026}, decomposition-based methods iteratively generated executable programs\cite{Binder-ICLR2023}, symbolic operations\cite{Chain-of-Table-ICLR2024}, or context-aware table partitions\cite{Dater-SIGIR2023} to break complex table reasoning into manageable steps, such as intermediate tables or sub-questions. Furthermore, some efforts exploited unified knowledge representations\cite{TrustUQA-AAAI2025} or explicit entity dependencies\cite{TIDE-ICLR2025} to enhance LLMs’ comprehension of table structures and question patterns. To harness the collaborative synergies of LLMs\cite{Reflexion-NeurIPS2023, Reflective-NeurIPS2024, AgentVerse-ICLR2024, Debate-ICML2024, GraphDiffusion-ICML2025}, recent multi-agent frameworks have integrated iterative reasoning-execution strategy\cite{ReAcTable-VLDB2024} and critique-refinement loops\cite{TableCritic-ACL2025} for agent collaboration, substantially improving the robustness and accuracy of table reasoning.

Although previous methods have achieved promising results, they typically focused on the overall assessment and linearized tables into long-context input, which not only disrupted the row-column-cell structure but also incurred the ``lost-in-the-middle'' issue\cite{Lost-in-the-middle-TACL2024}. Moreover, these approaches mainly relied on reasoning from scratch, without leveraging historical experiences to facilitate subsequent inference. To address these issues, we introduce the \emph{Operation-wise TableQA} task for fine-grained evaluation, and the proposed framework transforms tables into attributed graphs for structured reasoning, and further abstracts historical experiences into reusable skills to continually augment table graph reasoning with self-evolution.

\subsection{Self-Evolution for Large Language Model}
Self-evolution methods~\cite{Self-Evolution-AAAI2026} enhance LLM behavior by incorporating self-reflection, iterative revision, or feedback-driven refinement. Representative works explore this direction from diverse perspectives: Recursive Introspection\cite{RISE-NeurIPS2024} enables LLMs to iteratively revisit and refine prior reasoning attempts through repeated self-evaluation. Self-Evolving Curriculum\cite{SEC-arXiv2025} focuses on the automatic construction of curricula to progressively strengthen reasoning capabilities. In addition, Uncertainty-enhanced Preference Optimization\cite{UPO-AAAI2025} improves LLM policies by leveraging uncertainty-aware feedback sampling to yield more reliable learning signals.

Unlike prior self-evolution approaches that refine LLM behavior through model training or implicit policy shifts, our SkillTGR is complementary to externalize reusable problem-solving patterns into structured skills without any parameter updating. These skills are continuously evolved through explicit operations such as revision and merging, which makes the improvement process more transparent and controllable.

\section{Methodology}
In this section, we first formalize the \emph{Operation-wise TableQA} task and then elaborate on our SkillTGR framework.

\subsection{Task Definition}
TableQA aims to infer answers to natural language questions based on table context for overall assessment. Formally, given the column headers $\mathcal{H} = \{h_j \mid 1 \leq j \leq n\}$ and a table $\mathcal{T}=\{c_{ij}\mid 1\leq i\leq m, 1\leq j\leq n\}$ with $m$ rows and $n$ columns, $c_{ij}$ denotes as the cell value at the $i$-th row and $j$-th column. Together with the associated metadata $\mathcal{M}$ (e.g., table caption) and a natural language question $\mathcal{Q}$, the TableQA task requires models to optimize the mapping function $F(\mathcal{H}, \mathcal{T}, \mathcal{M}, \mathcal{Q})\rightarrow \mathcal{A}$ to generate an answer $\mathcal{A}$ that accurately responds to the question $\mathcal{Q}$. Accordingly, the answer $\mathcal{A}$ can correspond to a span of cell values, a set of entries, or an aggregated numerical result, depending on the semantic intent of the question $\mathcal{Q}$. Based on this, the \emph{Operation-wise TableQA} task associates each question $\mathcal{Q}$ with an operation type $ \mathcal{O}_\mathcal{Q}$ (i.e., \emph{Conditional Lookup}, \emph{Comparison}, \emph{Extremum}, \emph{Aggregation} or \emph{Arithmetic}) for operation-wise evaluation, enabling fine-grained diagnosis of model capabilities beyond overall performance.

\subsection{Overview}
\begin{figure*}[ht]
    \centering
    \includegraphics[width=1.0\linewidth]{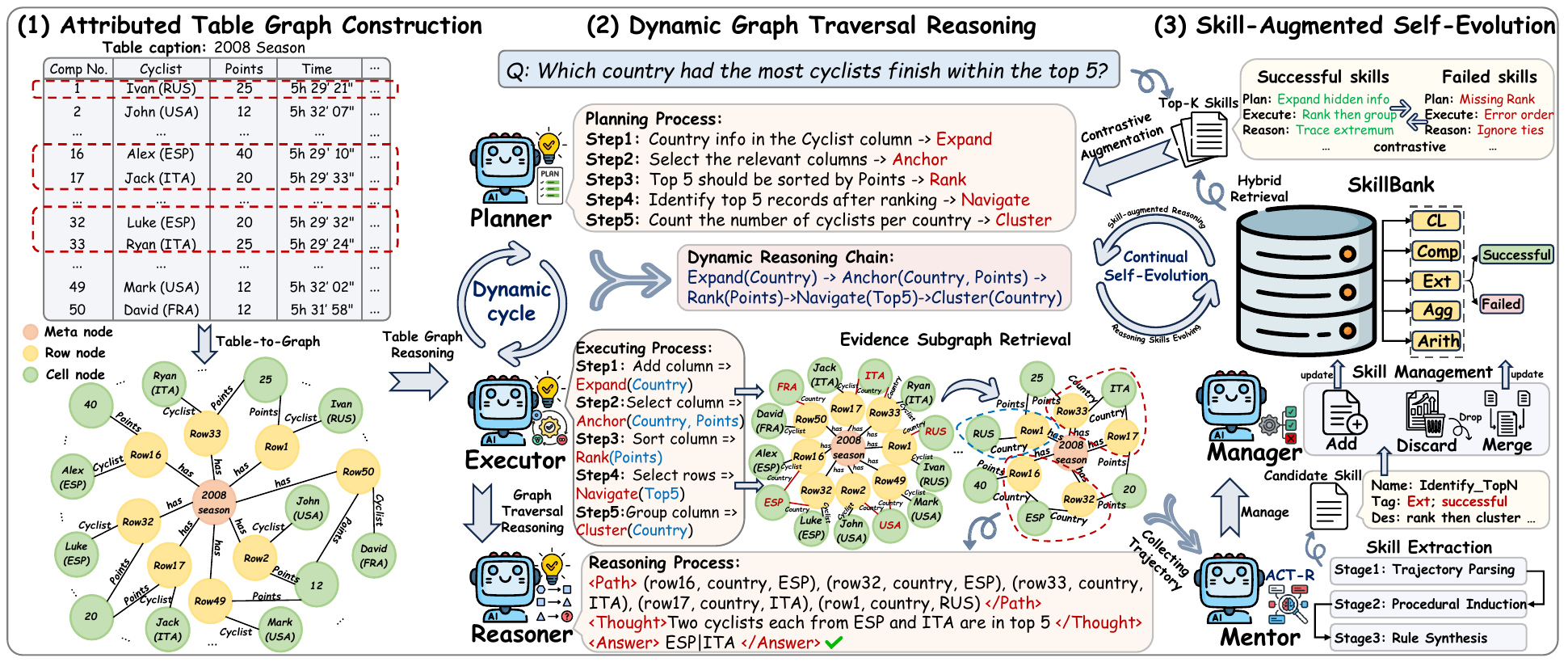}
    \caption{The overall architecture of our SkillTGR framework.}
    \label{fig:model}
\end{figure*}

To address the aforementioned bottlenecks, we propose SkillTGR, a skill-augmented table graph reasoning framework with three key components. (1) To overcome structural semantic loss from table linearization and mitigate the ``lost-in-the-middle'' issue, \textbf{Attributed Table Graph Construction} transforms tables into attributed graphs with explicit row-column-cell structures for cross-row relational modeling. (2) To facilitate structured evidence retrieval, \textbf{Dynamic Graph Traversal Reasoning} employs LLMs to plan and execute dynamic reasoning chains to retrieve evidence subgraphs for graph traversal reasoning. (3) To break the reason-from-scratch limitation and enable cross-instance experience transfer, \textbf{Skill-Augmented Self-Evolution} distills historical trajectories into reusable skills and performs contrastive augmented reasoning through both successful and failed skills, enabling continual self-evolution. The overview of SkillTGR is shown in Fig. \ref{fig:model}.

\subsection{Attributed Table Graph Construction}
As discussed above, existing methods typically linearize tables into flat long-context sequences for LLM inference. Such serialization inevitably disrupts the intrinsic row-column-cell structural semantics and weakens cross-row relational reasoning. Moreover, long sequential representations induce the ``lost-in-the-middle'' issue, causing LLMs to overlook critical intermediate evidence, limiting their effectiveness on multi-row retrieval and reasoning for complex operations.

To overcome these limitations, we represent the table as an attributed table graph to explicitly preserve the intrinsic row-column-cell structural semantics and establish relational dependencies among cross-row table elements.

Specifically, given a structured table $\mathcal{T}$ with $m$ rows and $n$ columns, together with its associated header $\mathcal{H}$ and metadata $\mathcal{M}$, we formalize these table content as an attributed table graph $\mathcal{G}=\{\mathcal{V}, \mathcal{E}\}$ with the explicit row-column-cell triple structures, where $\mathcal{V}$ and $\mathcal{E}$ denote the node set and edge set respectively. The graph construction process consists of the following two stages:

\noindent\textbf{Node Construction.} The node set $\mathcal{V}$ is composed of three hierarchically organized node types:
\begin{itemize}
    \item \textbf{Meta node $v_{meta}$:} A global anchor node represents the entire table $\mathcal{T}$, attributed with the table caption in metadata $\mathcal{M}$ as its semantic identifier, and serves as the entry point for graph traversal.
    
    \item \textbf{Row nodes $\{r_i\}_{i=1}^{m}$:} Each row node $r_i$ represents the $i$-th row of table $\mathcal{T}$, encoding the tuple-level semantics of a table record.

    \item \textbf{Cell nodes $\bigcup_{j=1}^{n} \bigcup_{k=1}^{|\mathcal{U}_j|} \{c_j^{(k)}\}$:} For each column $col_j$, we enumerate its distinct values and assign a unique cell node $c_j^{(k)}$ to the $k$-th unique value, where $\mathcal{U}_j$ denotes the set of unique values in $col_j$. Notably, identical values across different columns are treated as semantically independent nodes, thereby preserving column-specific semantics without cross-column merging.
\end{itemize}

\noindent\textbf{Edge Connection.} The edge set $\mathcal{E}$ encodes structural relationships at two levels:
\begin{itemize}
    \item \textbf{Meta-to-Row Edges:} Meta node $v_{meta}$ serves as the root node to connect all row nodes $\{r_i\}_{i=1}^{m}$ with ``has'' relation, thereby establishing a global structural backbone for top-down traversal.

    \item \textbf{Row-to-Cell Edges:} An undirected edge $(r_i, c_j^{(k)})$ is established if the value at position $(i, j)$ in table $\mathcal{T}$ corresponds to the unique value represented by $c_j^{(k)}$. Each edge is annotated with the column header $h_j$ as an attribute, enabling explicit traceability of the column-level provenance for each cell value.
\end{itemize}

Accordingly, the relational structure of table $\mathcal{T}$ can be decomposed into a set of structured triples at the cell level in the attributed table graph $\mathcal{G}$, defined as $\bigcup_{i=1}^{m} \bigcup_{j=1}^{n} \left\{ \langle r_i, h_j, c_{ij} \rangle \right\}$, where $c_{ij}$ denotes the cell value at row $r_i$ and column $col_j$. Triples sharing the same column $col_j$ and identical cell value $c_{ij}$ are connected to the same cell node $c_j^{(k)}$, enabling efficient value-anchored retrieval across multiple rows.

We utilize the open-source \emph{NetworkX}\footnote{https://networkx.org/en/} toolkit to transform tables into graphs in the offline manner. The graph construction over all $m \times n$ cell entries runs in $O(m \cdot n \cdot \log m)$ and takes $O(m \cdot n)$ space to store, where $\log m$ accounts for the time cost of checking whether a cell node is unique.

\subsection{Dynamic Graph Traversal Reasoning}
Based on the attributed table graph, our proposed SkillTGR framework assigns LLMs with distinct roles in a collaborative multi-agent system, namely the Planner, Executor, and Reasoner, to progressively perform operator-based dynamic chain planning, evidence subgraph retrieval, and graph traversal reasoning for answer derivation.

\begin{table}[ht]
    \centering
    \caption{Graph operators on attributed table graph}
    \begin{tabularx}{\columnwidth}{c|X}
    \toprule
    \textbf{Operator} & \multicolumn{1}{c}{\textbf{Description}} \\
    \midrule
    \raisebox{-6.5mm}{\emph{Expand}}
    & Identify composite cell values in a target column, decompose them into fine-grained attributes, and extend the graph with newly derived column edges and cell nodes, thereby enriching the table graph for subsequent precise retrieval, analogous to \emph{adding columns} in tabular processing. \\ \midrule
    \raisebox{-6.5mm}{\emph{Navigate}}
    & Locate condition-constrained column edges and traverse connected cell nodes to verify value constraints, then backtrack to associated row nodes and target cell nodes for qualifying evidence, thereby enabling condition-guided multi-hop graph traversal, analogous to \emph{selecting rows} in tabular processing. \\ \midrule
    \raisebox{-4.5mm}{\emph{Anchor}}
    & Match candidate column edges according to question semantics, and retrieve their associated row and cell nodes, thereby constructing a local evidence subgraph for subsequent reasoning, analogous to \emph{selecting columns} in tabular processing. \\ \midrule
    \raisebox{-4.5mm}{\emph{Cluster}}
    & Group cell nodes sharing the same semantics in target column edges into community-like structures, thereby facilitating aggregation-oriented reasoning, analogous to \emph{grouping} in tabular processing. \\ \midrule
    \raisebox{-4.5mm}{\emph{Rank}}
    & Sort cell nodes associated with target column edges by numerical values, enabling the extraction of top-k, maximum, or minimum nodes for comparative and extremum reasoning, analogous to \emph{sorting} in tabular processing. \\
    \bottomrule
    \end{tabularx}
    \label{tab:graph_operator}
\end{table}

Specifically, to enable LLMs to effectively plan, retrieve and reason over attributed table graphs, we define a set of LLM-friendly graph operator primitives $\mathcal{O}_\mathcal{G}$, including \emph{Expand}, \emph{Navigate}, \emph{Anchor}, \emph{Cluster} and \emph{Rank} described in Table~\ref{tab:graph_operator}.

Based on this, the LLM agents Planner and Executor work in a collaborative manner to retrieve the minimal yet sufficient evidence subgraph. The Planner focuses on high-level decision-making by selecting appropriate graph operators and organizing the reasoning chain. Meanwhile, the Executor translates these abstract decisions into concrete graph-level actions and executes the corresponding operators to update the evidence subgraph accordingly. Through iterative planning and execution, the evidence subgraph is progressively refined along the dynamic chain into a minimal yet sufficient structure for subsequent reasoning. 

At reasoning step $t$, let $\mathcal{C}_{<t} = \{(o_1, \mathbf{a}_1), \ldots, (o_{t-1}, \mathbf{a}_{t-1})\}$ denote the historical dynamic reasoning chain, where $o_i \in \mathcal{O}_\mathcal{G}$ represents the graph operator selected at step $i$ and $\mathbf{a}_i$ denotes its corresponding execution arguments. Given the question $\mathcal{Q}$, the historical reasoning chain $\mathcal{C}_{<t}$, and the current evidence subgraph $\mathcal{G}_{t-1}$ (initialized as  $\mathcal{G}_{0}=\mathcal{G}$), the Planner is instructed by the planning prompt $\mathcal{P}_{\text{plan}}$ to dynamically determine the next graph operator for the current reasoning step:
\begin{equation}
    o_t = \pi_\text{Planner}(\mathcal{Q},\ \mathcal{C}_{<t},\ \mathcal{G}_{t-1};\ \mathcal{P}_{\text{plan}}),
\end{equation}
where $o_t \in \mathcal{O}_\mathcal{G}$ denotes the reasoning action at step $t$.

Upon receiving the planned operator, the Executor generates the corresponding execution arguments conditioned on the execution prompt $\mathcal{P}_{\text{exec}}$, and subsequently executes the graph operator to retrieve the updated evidence subgraph:
\begin{equation}
    \mathbf{a}_t = \pi_\text{Executor}(\mathcal{Q},\ \mathcal{G}_{t-1}, \ o_t;\ \mathcal{P}_{\text{exec}}),
\end{equation}
\begin{equation}
    \mathcal{G}_t = o_t(\mathcal{G}_{t-1}, \ \mathbf{a}_t),
\end{equation}
where $\mathbf{a}_t$ denotes the execution arguments for operator $o_t$, and $\mathcal{G}_t$ represents the retrieved evidence subgraph at step $t$.

The aforementioned iterative interaction continues until the Planner emits a termination signal $\langle\text{EOS}\rangle$ or reaches the maximum step $T$, thereby yielding the final evidence subgraph $\mathcal{G}^{*} $, which can be formulated as:
\begin{equation}
    \mathcal{G}^* = \text{Execute}\left(\mathcal{G}_{0},\ \{(o_t, \mathbf{a}_t)\}_{t=1}^{T}\right),
\end{equation}
where each step in the reasoning chain $\mathcal{C} = \{(o_t, \mathbf{a}_t)\}_{t=1}^{T}$ is dynamically conditioned on preceding actions and the evolving subgraph, thereby enabling progressive and adaptive evidence retrieval in a structured and interpretable manner.

Accordingly, the LLM agent Reasoner performs structured graph traversal reasoning over the retrieved evidence subgraph to derive the final answer. Specifically, given the question $\mathcal{Q}$, the dynamic reasoning chain $\mathcal{C}$ with the accumulated procedural information (e.g., grouping statistics and ranking results), the Reasoner instructed by the reasoning prompt $\mathcal{P}_{\text{reason}}$ to conduct traversal-based reasoning over evidence subgraph $\mathcal{G}^*$ via Chain-of-Thought, generating the explicit structured reasoning path composed of row-column-cell triples and inferring the final answer. The reasoning process is formulated as:
\begin{equation}
     \mathcal{R},\ \mathrm{T},\ \mathcal{A} = \pi_\text{Reasoner}\bigl(\mathcal{Q},\ \mathcal{C},\ \mathcal{G}^*;\ \mathcal{P}_{\text{reason}}\bigr),
\end{equation}
where $\mathcal{R}$ denotes the reasoning path, $\mathrm{T}$ is the Chain-of-Thought text and $\mathcal{A}$ represents the derived answer. By progressively aggregating semantic evidence along the reasoning path, the Reasoner grounds each inferential decision in an explicit graph traversal trajectory, enhancing both interpretability and reliability in answer derivation. 

Although the above multi-agent collaborative framework enables dynamic structured reasoning over attributed table graphs, it still follows a reason-from-scratch paradigm, where agent capabilities remain stagnant and fail to leverage the historical reasoning experience. Therefore, this limitation motivates the incorporation of reusable reasoning skills for continual self-evolution and enhanced reasoning capability.

\subsection{Skill-Augmented Self-Evolution}
As discussed above, existing methods uniformly employ the reason-from-scratch paradigm for table reasoning, which is the modeling bottleneck to infer each question independently without leveraging historical reasoning experience. Consequently, reasoning patterns for structurally similar operation-wise questions can neither be effectively reused nor transferred across instances, imposing constraints on both reasoning efficiency and answer accuracy.

To address this limitation, our proposed SkillTGR framework introduces a plug-and-play hierarchical skill library SkillBank, which distills historical reasoning trajectories into compact reusable skills, thereby augmenting the multi-agent framework with the continual self-evolution capability for dynamic graph traversal reasoning.

\subsubsection{\textbf{Skill Extraction}}
Our SkillTGR first collects reasoning trajectories by rolling out the environment from dynamic graph traversal reasoning. For a given question $\mathcal{Q}$, the reasoning trajectory is formally defined as $\tau = \bigl(\mathcal{Q},\ \mathcal{C},\ \mathcal{G}^*,\ \mathcal{R},\ \mathrm{T},\ \mathcal{A}\bigr)$ as produced by the dynamic graph traversal reasoning process. However, raw trajectories are inherently verbose and instance-specific, precluding direct reuse across instances. Therefore, how to distill concise and reusable procedural knowledge from such redundant experience remains the pivotal challenge for enabling self-evolving reasoning.

Inspired by the \emph{Adaptive Control of Thought-Rational} (ACT-R) theory~\cite{ACT-R} in cognitive science, which models how humans progressively abstract declarative memory into procedural knowledge, our SkillTGR introduces the LLM agent Mentor to progressively distill the verbose reasoning trajectories into compact transferable skills through the following three-stage knowledge compilation process:

\noindent\textbf{Stage 1: Trajectory Parsing.} Given a reasoning trajectory $\tau$, the Mentor performs the structured analysis instructed by the parsing prompt $\mathcal{P}_{\text{parse}}$, to identify the reasoning objective, critical decision points, and dependencies among reasoning actions. In addition, the Mentor further critiques the trajectory to verify whether the reasoning process derives the correct answer, thereby categorizing it as a successful $\tau^{+}$ or failed trajectory $\tau^{-}$ for differential induction in the subsequent stage. The trajectory parsing is formulated as:
\begin{equation}
    \mathcal{D},\ \tau^{\sigma} = \pi_\text{Mentor}^{\text{parse}}\bigl(\tau;\ \mathcal{P}_{\text{parse}}\bigr), \quad \sigma \in \{+, -\},
\end{equation}
where $\mathcal{D}$ is the description of structured analysis, and $\tau^{\sigma}$ denotes the successful or failed trajectory.

\noindent\textbf{Stage 2: Procedural Induction.} Conditioned on the structured analysis and induction prompt $\mathcal{P}_{\text{induce}}$, the Mentor employs differential strategies to induce procedural knowledge from the labeled trajectory. For the successful trajectory $\tau^{+}$, it extracts the strategic patterns underlying correct reasoning and transferable across instances of the same operation type. For the failed trajectory $\tau^{-}$, it summarizes failure lessons that reveal the root causes of reasoning errors to avoid. The induced experiences are subsequently generalized into a set of instance-independent production rules:
\begin{equation}
    \mathcal{R}_{\text{prod}}^{\sigma} = \pi_\text{Mentor}^{\text{induce}}\bigl(\mathcal{D},\ \tau^{\sigma};\ \mathcal{P}_{\text{induce}}\bigr), \quad \sigma \in \{+, -\},
\end{equation}
where $\mathcal{R}_{\text{prod}}^{\sigma}$ denotes the production rules induced from the successful or failed reasoning trajectory.

\noindent\textbf{Stage 3: Rule Synthesis.} To consolidate these production rules into compact reasoning knowledge, the Mentor distills and synthesizes the operation-wise reasoning patterns or failure lessons into structured forms, yielding the high-level candidate skill instructed by the synthesis prompt $\mathcal{P}_{\text{synth}}$:
\begin{equation}
    s_{cand}^{\sigma} = \pi_\text{Mentor}^{\text{synth}}\bigl(\mathcal{Q},\ \mathcal{R}_{\text{prod}}^{\sigma};\ \mathcal{P}_{\text{synth}}\bigr), \quad \sigma \in \{+, -\},
\end{equation}
where $s_{cand}^{\sigma}$ is the candidate skill for subsequent management.

\begin{figure}
    \centering
    \includegraphics[width=1.0\linewidth]{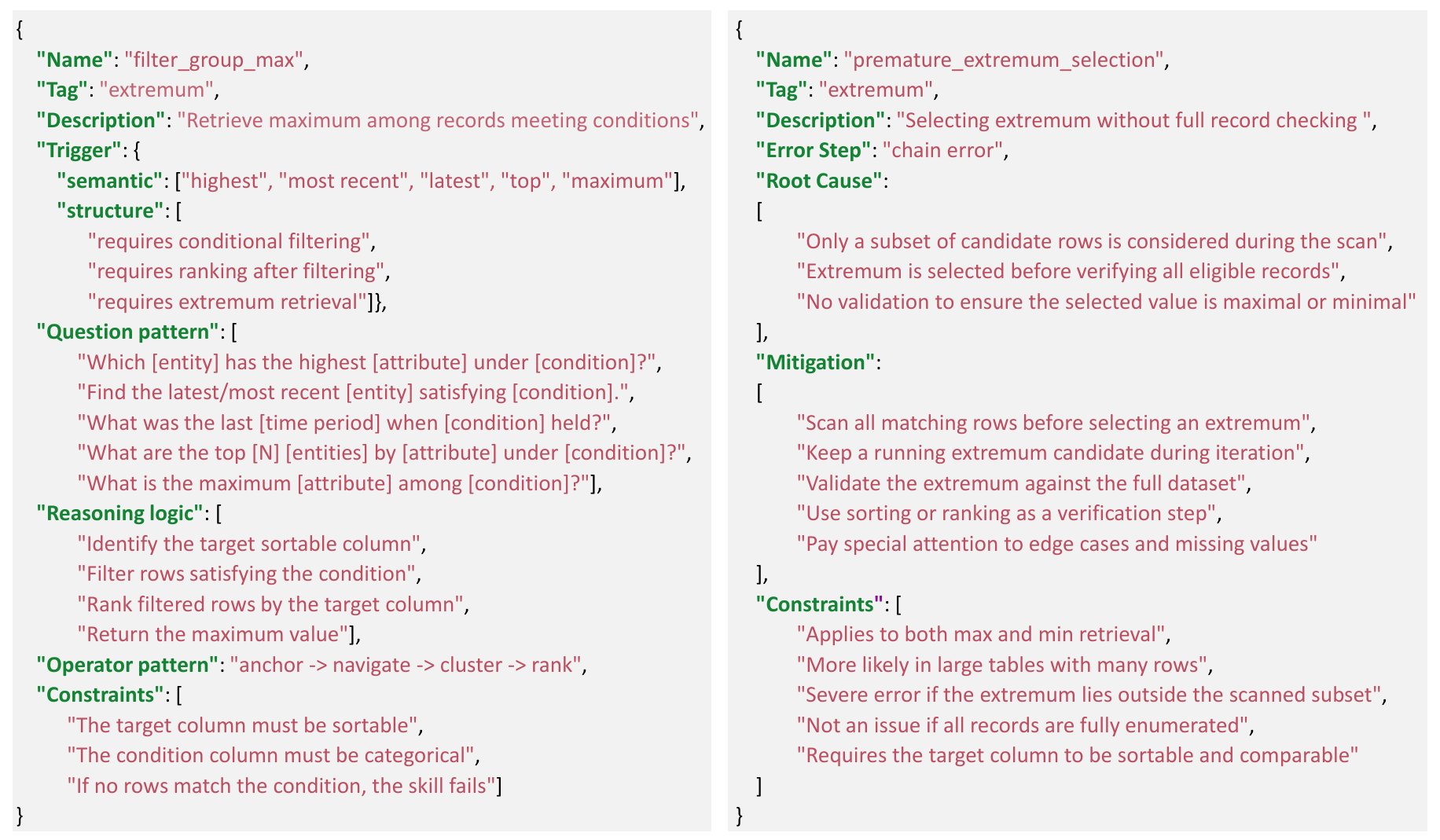}
    \caption{Examples of the successful (left) and failed (right) skills}
    \label{fig:skill_demo}
\end{figure}

Fig.~\ref{fig:skill_demo} illustrates the JSON-style examples of successful and failed skills. Through the above skill extraction under cognitive heuristics, the Mentor progressively distills the reasoning trajectories into transferable candidate skills for SkillBank construction and continual evolution.

\subsubsection{\textbf{Hierarchical SkillBank}}
Our SkillTGR framework organizes the distilled knowledge into a hierarchical skill library SkillBank, which enables efficient retrieval of operation-wise insights during dynamic graph traversal reasoning, and continuously evolves throughout the reasoning process to endow the multi-agent framework with self-evolution capability.

Specifically, SkillBank $\mathcal{B}$ is structured as a two-level hierarchy aligned with our proposed \emph{Operation-wise TableQA} task. At the first level, skills are partitioned according to operation-wise question taxonomy $\mathcal{O}_\mathcal{Q}$ (i.e., \emph{Conditional Lookup}, \emph{Comparison}, \emph{Extremum}, \emph{Aggregation} and \emph{Arithmetic}), thereby ensuring the operation-wise skill retrieval. At the second level, skills within each operation type are bifurcated by trajectory status $\sigma \in \{+, -\}$ into successful skills and failed skills, which encapsulate transferable reasoning strategies and instructive failure lessons respectively. Formally:
\begin{equation}
    \mathcal{B} = \bigcup\limits_{o \in \mathcal{O}_\mathcal{Q}} \Bigl( \mathcal{B}_{o}^{+} \cup \mathcal{B}_{o}^{-} \Bigr)
\end{equation}
Such a hierarchical architecture brings two key advantages: operation-wise partitioning eliminates cross-operation interference, while the dual-track design enables the LLM agent to simultaneously leverage successful strategies and failure heuristics for experience-augmented reasoning.

Prior to admission into SkillBank, each candidate skill $s_{cand}^{\sigma}$ is evaluated by the LLM agent Manager in terms of completeness, portability and compactness, yielding a decision $d \in \{Add,\ Discard,\ Merge\}$ for skill management. To support this evaluation, the Manager first employs hybrid retrieval to identify the most relevant skill $\hat{s}_{o}^{\sigma}$ from the corresponding SkillBank branch $\mathcal{B}_{o}^{\sigma}$, which shares the same operation type and skill status:
\begin{equation}
    \hat{s}_{o}^{\sigma} = \arg\max\limits_{s_i \in \mathcal{B}_{o}^{\sigma}}\ \Bigl[\alpha \cdot Sem(s_{cand}^{\sigma}, s_i) + (1-\alpha) \cdot Lex(s_{cand}^{\sigma}, s_i)\Bigr],
\end{equation}
where $\alpha$ denotes the weight to trade-off the dense semantic similarity\footnote{https://huggingface.co/sentence-transformers} and lexical BM25 matching for hybrid retrieval. Conditioned on candidate skill $s_{cand}^{\sigma}$ and its relevant skill $\hat{s}_{o}^{\sigma}$, the Manager then derives the decision via prompt $\mathcal{P}_{\text{manage}}$:
\begin{equation}
    d,\ s^* = \pi_{\text{Manager}}\bigl(s_{cand}^{\sigma},\ \hat{s}_{o}^{\sigma};\ \mathcal{P}_{\text{manage}}\bigr),
\end{equation}
where $d \in \{Add,\ Discard,\ Merge\}$ denotes the management decision and $s^*$ is the merged skill or $\varnothing$. Specifically, $Add$ incorporates the candidate skill $s_{cand}^{\sigma}$ as a new entry when it is complete and non-redundant. While $Discard$ drops it if knowledge-deficient or duplicative. $Merge$ consolidates the candidate skill $s_{cand}^{\sigma}$ and its most relevant skill $\hat{s}_{o}^{\sigma}$ into a refined skill $s^*$ when they are similar yet complementary. The evolution of SkillBank is formalized as:
\begin{equation}
    \mathcal{B}_{o}^{\sigma^*} = \begin{cases} \mathcal{B}_{o}^{\sigma} \cup \{s_{cand}^{\sigma}\}, & d = Add \\[4pt] \mathcal{B}_{o}^{\sigma}, & d = Discard \\[4pt] \bigl(\mathcal{B}_{o}^{\sigma} \setminus \{\hat{s}_{o}^{\sigma}\}\bigr) \cup \{s^*\}, & d = Merge \end{cases}
\end{equation}

Through the synergy of hierarchical architecture and management decision, SkillBank continuously accumulates and refines operation-wise skills throughout the reasoning process, forming the self-evolution loop that progressively strengthens experiential support for dynamic graph traversal reasoning.

\subsubsection{\textbf{Skill-augmented Reasoning}}
Upon constructing the hierarchical SkillBank, our SkillTGR introduces the contrastive augmentation mechanism that incorporates both successful and failed skills into each pivotal step of dynamic graph traversal reasoning, achieving coordinated enhancement of the Planner, Executor and Reasoner.

Specifically, given a question $\mathcal{Q}$ with its identified operation type $o \in \mathcal{O}_\mathcal{Q}$, our SkillTGR first retrieves top-k successful and failed skill candidates from the corresponding SkillBank branches $\mathcal{B}_{o}^{\sigma}$ via hybrid retrieval, and then filters them by the similarity threshold $\eta$ to form a contrastive skill set:
\begin{equation}
    \hat{S}_{o}^{\sigma} = \left\{ s_i \in \operatorname{Top-k}\!\left(\mathcal{B}_{o}^{\sigma};\ k\right) \;\middle|\; \text{Score}(\mathcal{Q}, s_i) \geq \eta \right\},
\end{equation}
\begin{equation}
    \text{Score}(\mathcal{Q}, s_i) = \alpha \cdot \text{Sem}(\mathcal{Q}, s_i) + (1-\alpha) \cdot \text{Lex}(\mathcal{Q}, s_i),
\end{equation}
where $\hat{S}_{o}^{\sigma} = \langle \hat{S}_{o}^{+},\ \hat{S}_{o}^{-} \rangle$ denotes the contrastive skill set consisting of $k$ successful skills and $k$ failed skills, serving as the experiential prior injected into the Planner, Executor and Reasoner to augment their reasoning processes.

\noindent\textbf{Planner Augmentation.} The Planner is responsible for dynamically selecting the appropriate graph operator $o_t \in \mathcal{O}_\mathcal{G}$ at each reasoning step $t$. Without explicit experiential guidance, the Planner is prone to redundant exploration or erroneous operator selection when reasoning from scratch.

By incorporating contrastive skill augmentation, successful skills $\hat{S}_{o}^{+}$ supply validated operator chain patterns as a positive anchor that steers the Planner toward high-confidence chain and accelerates planning convergence, while failed skills $\hat{S}_{o}^{-}$ expose historical planning pitfalls as the negative constraint to suppress erroneous chain recurrence. The skill-augmented Planner is formalized as:
\begin{equation}
    o_t = \pi_\text{Planner}\!\left(\mathcal{Q},\ \mathcal{C}_{<t},\ \mathcal{G}_{t-1},\ \hat{S}_{o}^{+},\ \hat{S}_{o}^{-};\ \mathcal{P}_{\text{plan}}^{\text{skill}}\right),
\end{equation}
Where $\mathcal{P}_{\text{plan}}^{\text{skill}}$ denotes the planning prompt augmented with contrastive skills. The contrastive guidance orients the Planner toward successful planning patterns while avoiding failure-prone ones, thereby jointly improving planning accuracy and reducing search overhead. 

\noindent\textbf{Executor Augmentation.} The Executor generates precise execution arguments $\mathbf{a}_t$ conditioned on the operator $o_t$ and updates the evidence subgraph $\mathcal{G}_t$. The quality of argument generation directly determines the fidelity of evidence subgraph retrieval, with argument errors constituting the primary source of subgraph deviation.

\begin{algorithm}[t]
\caption{The overall workflow of SkillTGR}
\label{alg:SkillTGR}

\textbf{Input:} Question set $\mathcal{Q}_{set}$, tables $\mathcal{T}$ \\
\textbf{Output:} Answer set $\mathcal{A}_{set}$, evolved SkillBank $\mathcal{B}^*$
\begin{algorithmic}[1]
\State Initialize SkillBank $\mathcal{B} \leftarrow \bigcup_{o \in \mathcal{O}_\mathcal{Q}} \bigl(\mathcal{B}_o^{+} \cup \mathcal{B}_o^{-}\bigr)$

\State \textcolor{blue}{\textit{// Attributed Table Graph Construction}}
    
\State $\mathcal{G} \leftarrow \text{Table2Graph}(\mathcal{T})$ \hfill $\triangleright$ \text{represent $\mathcal{T}$ as structured triples}

\For{$\mathcal{Q} \in \mathcal{Q}_{set}$}

    \State \textcolor{blue}{\textit{// Dynamic Graph Traversal Reasoning}}

    \State $\hat{S}_o^{\sigma} \leftarrow \text{SkillRetrieve}(\mathcal{Q},\ \mathcal{B})$ \hfill $\triangleright$ \text{Eq.~(13)--(14)}

    \State $\mathcal{C} \leftarrow \text{Planner}(\mathcal{Q},\ \mathcal{G},\ \hat{S}_o^{\sigma})$ \hfill $\triangleright$ \text{Eq.~(15)}

    \State $\mathcal{G}^* \leftarrow \text{Executor}(\mathcal{Q},\ \mathcal{G},\ \mathcal{C},\ \hat{S}_o^{\sigma})$ \hfill $\triangleright$ \text{Eq.~(16)--(17)}

    \State $\mathcal{R},\ \mathrm{T},\ \mathcal{A} \leftarrow \text{Reasoner}(\mathcal{Q},\ \mathcal{G}^*,\ \mathcal{C},\ \hat{S}_o^{\sigma})$ \hfill $\triangleright$ \text{Eq.~(18)}

    \State $\mathcal{A}_{set} \leftarrow \mathcal{A}_{set} \cup \{\mathcal{A}\}$

    \State \textcolor{blue}{\textit{// Skill-Augmented Self-Evolution}}

    \State $\tau \leftarrow (\mathcal{Q},\ \mathcal{C},\ \mathcal{G}^*,\ \mathcal{R},\ \mathrm{T},\ \mathcal{A})$ \hfill $\triangleright$ \text{collect trajectory}

    \State $s_{cand}^{\sigma} \leftarrow \text{Mentor}(\tau)$ \hfill $\triangleright$ \text{Eq.~(6)--(8)}

    \State $\hat{s}_o^{\sigma} \leftarrow \text{SkillRetrieve}(s_{cand}^{\sigma},\ \mathcal{B})$ \hfill $\triangleright$ \text{Eq.~(10)}

    \State $d,\ s^* \leftarrow \text{Manager}(s_{cand}^{\sigma},\ \hat{s}_o^{\sigma})$ \hfill $\triangleright$ \text{Eq.~(11)}

    \State $\mathcal{B}^* \leftarrow \text{SkillBank}(s_{cand}^{\sigma},\ d,\ s^*)$  \hfill $\triangleright$ \text{Eq.~(12)}

\EndFor

\State \Return $\mathcal{A}_{set},\ \mathcal{B}^*$

\end{algorithmic}
\end{algorithm}

Building upon contrastive skill augmentation, successful skills $\hat{S}_{o}^{+}$ provide the Executor with successful argument-setting paradigms as positive exemplars to guide graph-aligned argument generation, while failed skills $\hat{S}_{o}^{-}$ record argument errors as negative warnings constraining the generation process. The skill-augmented Executor is formulated as:
\begin{equation}
    \mathbf{a}_t = \pi_\text{Executor}\!\left(\mathcal{Q},\ \mathcal{G}_{t-1},\ o_t,\ \hat{S}_{o}^{+},\ \hat{S}_{o}^{-};\ \mathcal{P}_{\text{exec}}^{\text{skill}}\right),
\end{equation}
\begin{equation}
    \mathcal{G}_t = o_t\!\left(\mathcal{G}_{t-1},\ \mathbf{a}_t\right),
\end{equation}
where $\mathcal{P}_{\text{exec}}^{\text{skill}}$ denotes the execution prompt augmented with contrastive skills. The mechanism enables the Executor to generate graph-aligned execution arguments within a well-constrained parameter space, thereby enhancing both the precision and completeness of evidence subgraph retrieval.

\noindent\textbf{Reasoner Augmentation.} The Reasoner conducts structured graph traversal reasoning over the final evidence subgraph $\mathcal{G}^*$ to generate an explicit reasoning path $\mathcal{R}$ and derive the final answer $\mathcal{A}$. The key challenge lies in generating coherent and reliable reasoning paths over evidence subgraphs while minimizing hallucinations and semantic errors.

Through contrastive skill augmentation, successful skills $\hat{S}_{o}^{+}$ provide validated traversal strategies as positive guidance to direct the Reasoner toward proven reasoning patterns and mitigate hallucinations, while failed skills $\hat{S}_{o}^{-}$ encode root causes of semantic misjudgments as negative constraints to reduce errors. The skill-augmented Reasoner is defined as:
\begin{equation}
    \mathcal{R},\ \mathrm{T},\ \mathcal{A} = \pi_\text{Reasoner}\!\left(\mathcal{Q},\ \mathcal{C},\ \mathcal{G}^*,\ \hat{S}_{o}^{+},\ \hat{S}_{o}^{-};\ \mathcal{P}_{\text{reason}}^{\text{skill}}\right),
\end{equation}
where $\mathcal{P}_{\text{reason}}^{\text{skill}}$ denotes the reasoning prompt augmented with contrastive skills. The contrastive augmentation mechanism facilitates the Reasoner toward accurate graph traversal and reliable answer generation.

In summary, successful and failed skills collectively establish an implicit contrastive augmentation mechanism for skill-augmented reasoning. By continuously distilling reasoning trajectories into transferable skills to enhance subsequent reasoning, SkillTGR realizes the continual self-evolution that progressively improves reasoning effectiveness through experience accumulation. The workflow of SkillTGR is summarized in Algorithm~\ref{alg:SkillTGR}, and prompt details are available in our codes.


\section{Experiments}

\subsection{Experimental Settings}
\begin{figure}[ht]
    \centering
    \begin{subfigure}[b]{0.48\linewidth}
        \centering
        \includegraphics[width=\linewidth]{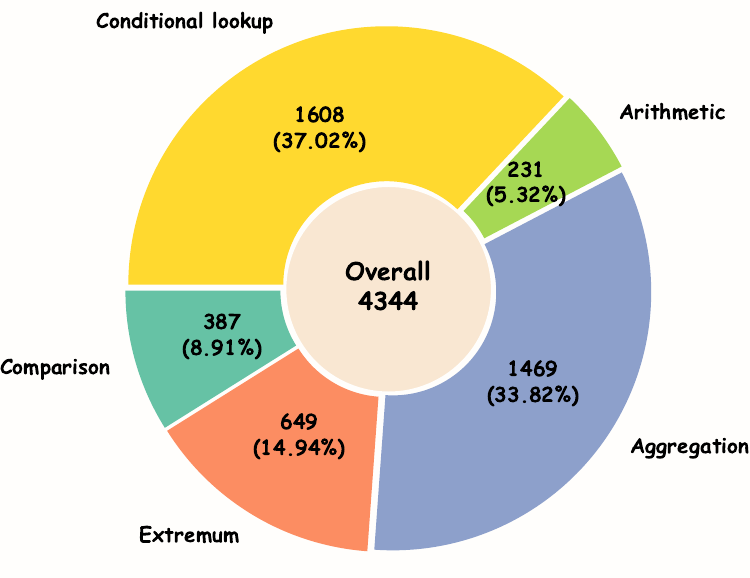}
        \caption{WikiTQ-ow}
        \label{fig:wikitq}
    \end{subfigure}
    \hfill
    \begin{subfigure}[b]{0.48\linewidth}
        \centering
        \includegraphics[width=\linewidth]{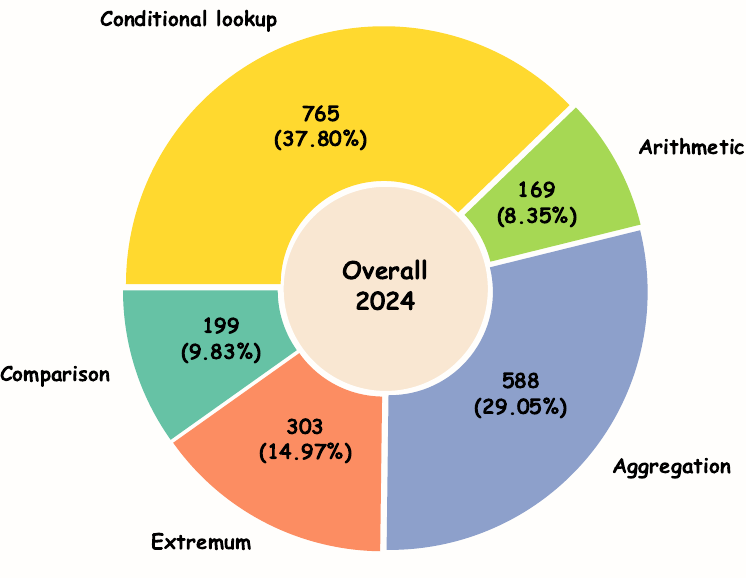}
        \caption{TabFact-ow}
        \label{fig:tabfact}
    \end{subfigure}
    \caption{Statistics of two operation-wise datasets.}
    \label{fig:dataset}
\end{figure}

\noindent\textbf{Datasets.} To explore the proposed \emph{Operation-wise TableQA} task, we introduce two benchmark datasets named WikiTQ-ow and TabFact-ow, which are refined by the widely recognized TableQA datasets under the operation-oriented taxonomy with five core reasoning types. Specifically, the \emph{Conditional Lookup (CL)} focuses on cell-level retrieval under one or multiple constraints. \emph{Comparison (Comp)}, \emph{Extremum (Ext)} and \emph{Aggregation (Agg)} involve local or global comparison, ranking, and statistical reasoning across multiple rows, while the \emph{Arithmetic (Arith)} further requires multi-step numerical computation. Fig.~\ref{fig:dataset} presents the statistics of both datasets, and the details of each dataset are described as follows:

\begin{itemize}
    \item \textbf{WikiTQ-ow:} WikiTQ\cite{WikiTQ-ACL2015} is a widely used table reasoning benchmark containing 4,344 test samples derived from 421 tables. To support the proposed \emph{Operation-wise TableQA} task, we refine it by employing the powerful GPT-5\cite{GPT-5-arXiv2025} to categorize questions into the above operation types with table contexts prompting. The annotations are verified by three domain postgraduates via majority voting, while cases with inconsistent judgments are further resolved by a senior doctoral expert.
    
    \item \textbf{TabFact-ow:} TabFact\cite{TabFact-ICLR2020} is a fact verification benchmark with 2,024 test samples from 298 tables. However, it exhibits two key limitations: (1) The binary answer space (entailed or refuted) allows models to predict correctly with erroneous evidence retrieval. (2) Its positive and negative samples are mainly constructed by modifying the core claim, repeatedly testing similar reasoning patterns. These issues restrict reasoning diversity and provide no explicit supervision over the underlying operations, making it insufficient for operation-wise evaluation. Therefore, we reformulate TabFact into QA forms following the WikiTQ paradigm. Specifically, GPT-5 is utilized to convert positive samples into QA pairs annotated with predefined operation types, and further generate diverse operation types of QA pairs to replace negative samples based on table contexts. The QA pairs are screened and verified by three domain postgraduates to enrich the dataset diversity and coverage. The quality control procedure remains consistent with WikiTQ-ow.
\end{itemize}

\noindent\textbf{Metric.} We follow previous studies to adopt the denotation accuracy as the evaluation metric, and report both the \textbf{\emph{overall}} and \textbf{\emph{operation-wise}} results across all experiments.

\noindent\textbf{Baselines.} To conduct the systematic evaluation, we perform comprehensive experiments comparing our SkillTGR against the following four categories of mainstream baselines: \textbf{(1) \emph{Generic Reasoning}:} E2E TQA, Few-Shot TQA~\cite{Fewshot-NeurIPS2020} and CoT TQA~\cite{CoT-NeurIPS2022}. \textbf{(2) \emph{Decomposition-based Reasoning}:} Binder\cite{Binder-ICLR2023}, Dater\cite{Dater-SIGIR2023} and Chain-of-Table\cite{Chain-of-Table-ICLR2024}. \textbf{(3) \emph{Structure-aware Reasoning}:} TrustUQA\cite{TrustUQA-AAAI2025} and TIDE\cite{TIDE-ICLR2025}. \textbf{(4) \emph{Agent-based Reasoning}:} ReAcTable\cite{ReAcTable-VLDB2024} and Table-Critic\cite{TableCritic-ACL2025}. Detailed descriptions of these baselines have been discussed in the \textsc{Introduction} and \textsc{Related Work} sections.

\noindent\textbf{Implementation Details.} To ensure comprehensive evaluation, we conduct experiments employing Qwen3.5-9B\footnote{https://huggingface.co/Qwen/Qwen3.5-9B} and Ministral3-8B\footnote{https://huggingface.co/mistralai/Ministral-3-8B-Instruct-2512} as backbone LLMs, which are deployed as API servers through the vLLM\cite{vLLM-SOSP2023} inference engine on a single NVIDIA L20-48G GPU with CUDA 12.4. For generic reasoning methods, we reproduce them using the prompt settings adopted in Chain-of-Table\cite{Chain-of-Table-ICLR2024}, while implementing other baselines with their default configurations to ensure optimal performance. For our SkillTGR, we warm up the SkillBank with 1,054 successful and 518 failed skills, which are distilled from 14,151 reasoning trajectories on the WikiTQ training set by skill management. For skill retrieval, we set the weight $\alpha$ as 0.8 to trade-off semantic and lexical matching, and retain the top-3 (i.e., k=3) retrieved skills whose matching scores exceed the threshold $\eta$ of 0.7. Across all experiments, we configure the temperature to 0.0 for greedy decoding.

\begin{table*}[ht]
\centering
\caption{Experimental results of \textbf{WikiTQ-ow} dataset. The best results are in \textbf{bold} and the second best are \underline{underlined}.}
\label{WikiTQ-result}
\scalebox{1.025}{
\begin{tabular}{lcccccc|cccccc}
\toprule
\multicolumn{1}{c}{\multirow{2}{*}[-0.8ex]{\textbf{Method}}} & \multicolumn{6}{c|}{\textbf{Qwen3.5-9B}}                                                      & \multicolumn{6}{c}{\textbf{Ministral3-8B}}                                     \\ \cmidrule{2-13} 
                                & \textbf{Overall} & \textbf{CL} & \textbf{Comp} & \textbf{Ext} & \textbf{Agg} & \textbf{Arith} & \textbf{Overall} & \textbf{CL} & \textbf{Comp} & \textbf{Ext} & \textbf{Agg} & \textbf{Arith} \\ \midrule
\rowcolor{gray!20} \multicolumn{13}{l}{\textbf{\emph{Generic Reasoning}}} \\
E2E TQA                         & 51.73            & 67.41       & 48.06         & 57.16        & 36.22        & 32.03          & 38.86            & 56.03       & 38.50         & 18.80        & 28.18        & 44.16          \\
Few-Shot TQA                    & 54.12            & 70.46       & 53.75         & 59.48        & 36.96        & 35.06          & 50.92            & 68.84       & 53.75         & 49.92        & 31.65        & 46.75          \\
CoT TQA                         & 63.56            & 73.82       & 61.50         & 62.56        & 53.91        & 59.74          &\underline{62.96}            & \underline{75.62}       & \underline{64.08}         & \underline{68.41}        & 47.11        & \underline{58.44}          \\ \midrule
\rowcolor{gray!20} \multicolumn{13}{l}{\textbf{\emph{Decomposition-based Reasoning}}} \\
Binder                          & 56.61            & 69.22       & 47.55         & 45.61        & 53.23        & 36.36          & 51.96            & 62.94       & 32.82         & 39.14        & 55.28        & 22.51          \\
Dater                           & 62.25            & 71.70       & 61.76         & 62.10        & 52.14        & 61.90          & 54.42            & 58.46       & 52.20         & 59.17        & 48.94        & 51.52          \\
Chain-of-Table                  & 66.83            & 74.01       & 65.37         & 63.48        & 61.33        & 63.64          & 55.64            & 66.54       & 55.81         & 54.39        & 46.02        & 44.16          \\ \midrule
\rowcolor{gray!20} \multicolumn{13}{l}{\textbf{\emph{Structure-aware Reasoning}}} \\
TrustUQA                        & 51.06            & 62.62       & 56.33         & 42.99        & 40.64        & 50.65          & 42.56            & 54.04       & 50.65         & 32.82        & 32.54        & 40.26          \\
TIDE                            & 63.74            & 71.89       & 62.27         & 65.49        & 54.60        & 62.77          & 55.34            & 61.82       & 50.39         & 51.31        & 52.08        & 50.65          \\ \midrule
\rowcolor{gray!20} \multicolumn{13}{l}{\textbf{\emph{Agent-based Reasoning}}} \\
ReAcTable                       & 60.57            & 70.40       & 58.91         & 57.63        & 52.55        & 54.11          & 56.31            & 66.67       & 57.88         & 52.08        & 48.06        & 45.89          \\
Table-Critic                    &\underline{79.42}	&\textbf{82.96}	&\underline{77.26}	&\underline{76.12}	&\underline{78.56}	&\underline{73.16}	&61.30	&67.10	&59.69	&61.94	&\underline{56.30}	&53.68          \\ \midrule
\rowcolor{gray!20} \multicolumn{13}{l}{\textbf{\emph{Skill-augmented Reasoning}}} \\
SkillTGR (Ours)                            &\textbf{80.39}	&\underline{82.34} &\textbf{79.84}	&\textbf{77.20} &\textbf{80.74}	&\textbf{74.46} &\textbf{72.97}	&\textbf{75.75} &\textbf{75.19}	&\textbf{71.96} &\textbf{70.80}	&\textbf{66.67}                \\ 
\multicolumn{1}{c}{$\triangle$}                            &\textcolor{red}{$\uparrow$0.97}	&\textcolor{teal}{$\downarrow$0.62}	&\textcolor{red}{$\uparrow$2.58}	&\textcolor{red}{$\uparrow$1.08}	&\textcolor{red}{$\uparrow$2.18}	&\textcolor{red}{$\uparrow$1.3}	&\textcolor{red}{$\uparrow$10.01}	&\textcolor{red}{$\uparrow$0.13}	&\textcolor{red}{$\uparrow$11.11}	&\textcolor{red}{$\uparrow$3.55}	&\textcolor{red}{$\uparrow$14.50}	&\textcolor{red}{$\uparrow$8.23}                \\ 
\bottomrule
\end{tabular}
}
\end{table*}

\begin{table*}[ht]
\centering
\caption{Experimental results of \textbf{TabFact-ow} dataset. The best results are in \textbf{bold} and the second best are \underline{underlined}.}
\label{TabFact-result}
\scalebox{1.025}{
\begin{tabular}{lcccccc|cccccc}
\toprule
\multicolumn{1}{c}{\multirow{2}{*}[-0.8ex]{\textbf{Method}}} & \multicolumn{6}{c|}{\textbf{Qwen3.5-9B}}                                                      & \multicolumn{6}{c}{\textbf{Ministral3-8B}}                                     \\ \cmidrule{2-13} 
                                & \textbf{Overall} & \textbf{CL} & \textbf{Comp} & \textbf{Ext} & \textbf{Agg} & \textbf{Arith} & \textbf{Overall} & \textbf{CL} & \textbf{Comp} & \textbf{Ext} & \textbf{Agg} & \textbf{Arith} \\ \midrule
\rowcolor{gray!20} \multicolumn{13}{l}{\textbf{\emph{Generic Reasoning}}} \\
E2E TQA                         &52.77	&71.63	&54.77	&51.82	&33.84	&32.54	&41.50	&65.36	&39.20	&20.46	&24.32	&33.73          \\
Few-Shot TQA                    &59.09	&72.81	&59.30	&54.13	&47.45	&46.15	&51.38	&70.85	&60.30	&50.83	&25.34	&44.38          \\
CoT TQA                         &67.84	&73.33	&67.34	&65.35	&63.10	&64.50	&\underline{65.37}	&\underline{70.98}	&66.83	&\underline{66.67}	&56.80	&\underline{65.68}          \\ \midrule
\rowcolor{gray!20} \multicolumn{13}{l}{\textbf{\emph{Decomposition-based Reasoning}}} \\
Binder                          &62.15	&71.76	&54.27	&53.14	&61.05	&47.93	&52.87	&64.31	&43.72	&41.58	&53.91	&28.40          \\
Dater                           &66.21	&72.68	&66.33	&63.37	&60.54	&61.54	&53.56	&57.52	&54.77	&55.12	&48.30	&49.70          \\
Chain-of-Table                  &73.22	&76.34	&72.86	&67.66	&72.79	&71.01	&54.59	&65.10	&62.81	&53.80	&44.73	&33.14          \\ \midrule
\rowcolor{gray!20} \multicolumn{13}{l}{\textbf{\emph{Structure-aware Reasoning}}} \\
TrustUQA                        &56.52	&68.50	&61.31	&49.83	&43.54	&53.85	&47.23	&58.56	&53.77	&40.26	&34.18	&46.15          \\
TIDE                            &68.77	&73.07	&69.85	&66.34	&64.97	&65.68	&61.66	&65.49	&61.31	&59.74	&59.35	&56.21          \\ \midrule
\rowcolor{gray!20} \multicolumn{13}{l}{\textbf{\emph{Agent-based Reasoning}}} \\
ReAcTable                       &65.02	&71.63	&65.83	&60.40	&59.86	&60.36	&58.60	&64.84	&61.31	&58.75	&51.36	&52.07          \\
Table-Critic                    &\underline{83.10}	&\underline{85.10}	&\underline{83.92}	&\underline{77.56}	&\underline{84.01}	&\underline{79.88}	&63.54	&63.53	&\underline{73.87}	&61.06	&\underline{62.59}	&59.17          \\ \midrule
\rowcolor{gray!20} \multicolumn{13}{l}{\textbf{\emph{Skill-augmented Reasoning}}} \\
SkillTGR (Ours)                            &\textbf{86.17}	&\textbf{86.27} &\textbf{88.44}	&\textbf{79.87} &\textbf{89.29}	&\textbf{83.43} &\textbf{74.95}	&\textbf{75.03} &\textbf{83.42}	&\textbf{72.61} &\textbf{72.79}	&\textbf{76.33}    \\ 
\multicolumn{1}{c}{$\triangle$}                            &\textcolor{red}{$\uparrow$3.07}	&\textcolor{red}{$\uparrow$1.17}	&\textcolor{red}{$\uparrow$4.52}	&\textcolor{red}{$\uparrow$2.31}	&\textcolor{red}{$\uparrow$5.28}	&\textcolor{red}{$\uparrow$3.55}	&\textcolor{red}{$\uparrow$9.58}	&\textcolor{red}{$\uparrow$4.05}	&\textcolor{red}{$\uparrow$9.55}	&\textcolor{red}{$\uparrow$5.94}	&\textcolor{red}{$\uparrow$10.20}	&\textcolor{red}{$\uparrow$10.65}                \\ 
\bottomrule
\end{tabular}
}
\end{table*}

\subsection{Main Results}
Table \ref{WikiTQ-result} and Table \ref{TabFact-result} report the main results across different LLMs on both datasets. Our systematic evaluation reveals several noteworthy observations as follows.

\textbf{First}, SkillTGR consistently outperforms all baseline methods both in overall and operation-wise assessment. On average, our method achieves 76.68\% overall accuracy on WikiTQ-ow and 80.56\% on TabFact-ow, yielding significant improvements of 5.49\% and 6.33\% over the strongest baselines respectively. Under the operation-wise evaluation, SkillTGR further brings average gains of 4.40\% and 5.72\% accuracy, highlighting its superiority in fine-grained table reasoning.

\textbf{Second}, most baselines exhibit a pronounced operation-wise performance gap: while performing well on simple \emph{Conditional Lookup} questions, they suffer notable degradation on other complex operations requiring cross-row reasoning and multi-step computation. Such limitations mainly stem from table linearization, which disrupts table structural semantics and induces the ``lost-in-the-middle'' issue in long contexts. In contrast, our SkillTGR models tables as attributed graphs with explicit row-column-cell structures for dynamic graph traversal reasoning, while leveraging reusable skills to transfer reasoning patterns across analogous instances. Consequently, our method maintains more balanced reasoning capabilities across diverse operation types and consistently achieves the best performance on complex operations, yielding significant improvements of 5.57\% and 6.50\% accuracy on average.

\begin{figure*}[ht]
\centering
\scalebox{0.87}{
\begin{subfigure}[t]{0.49\textwidth}
    \centering
    \includegraphics[width=\linewidth]{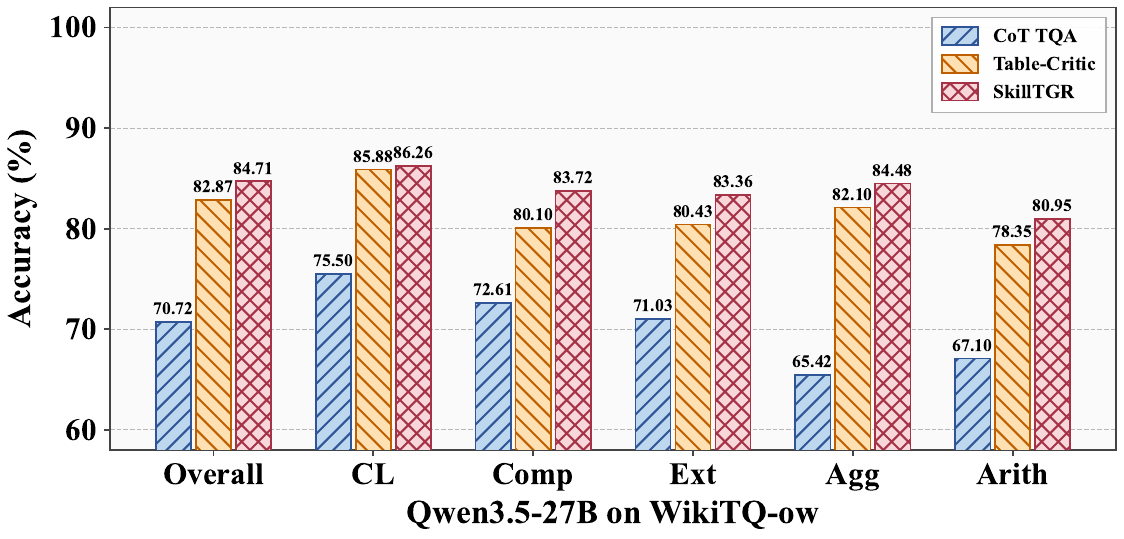}
    \includegraphics[width=\linewidth]{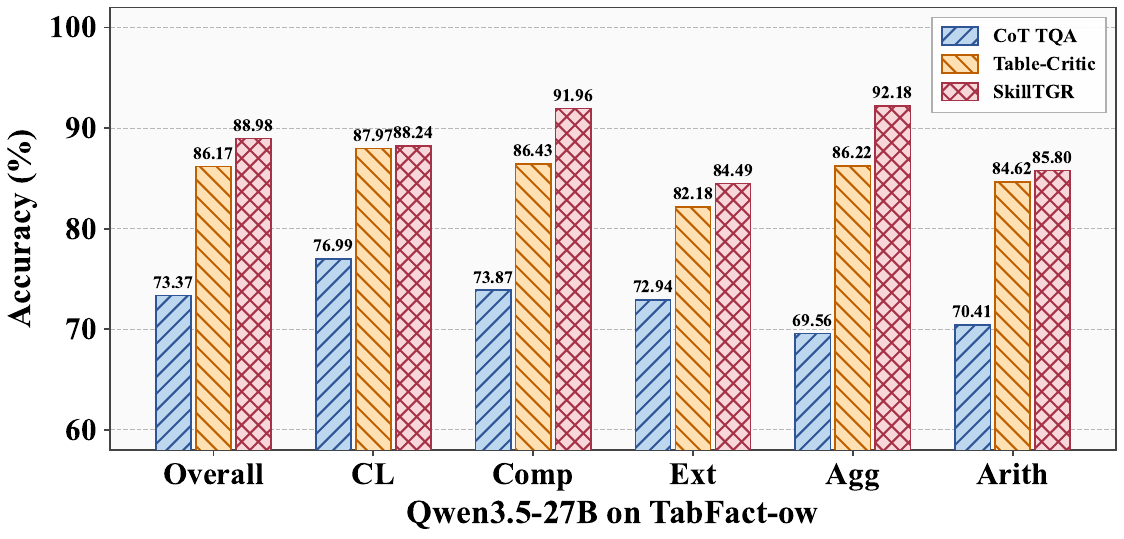}
    \label{fig:scaling_qwen}
\end{subfigure}
\hfill
\begin{subfigure}[t]{0.49\textwidth}
    \centering
    \includegraphics[width=\linewidth]{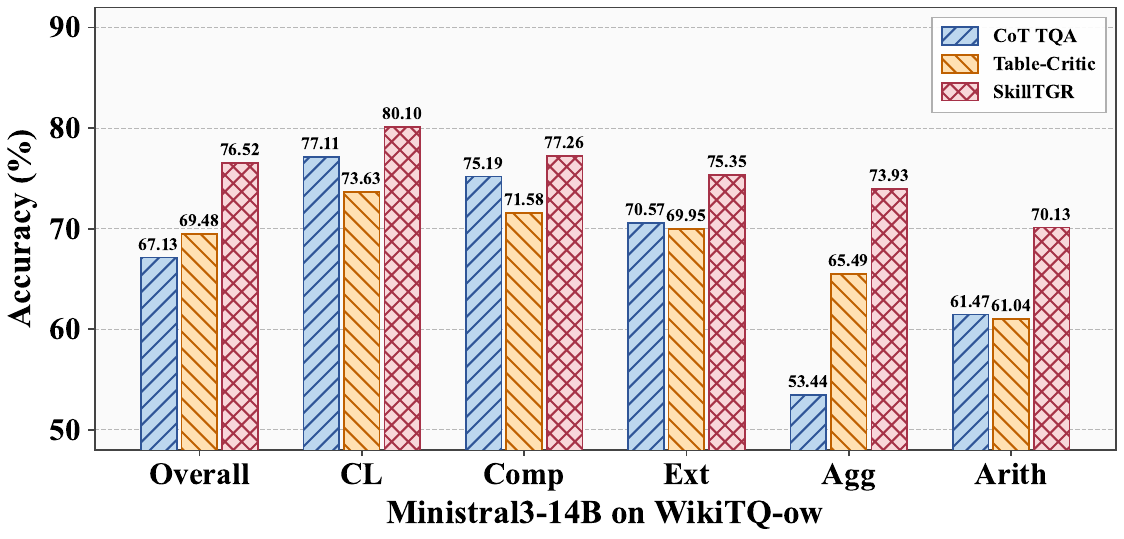}
    \includegraphics[width=\linewidth]{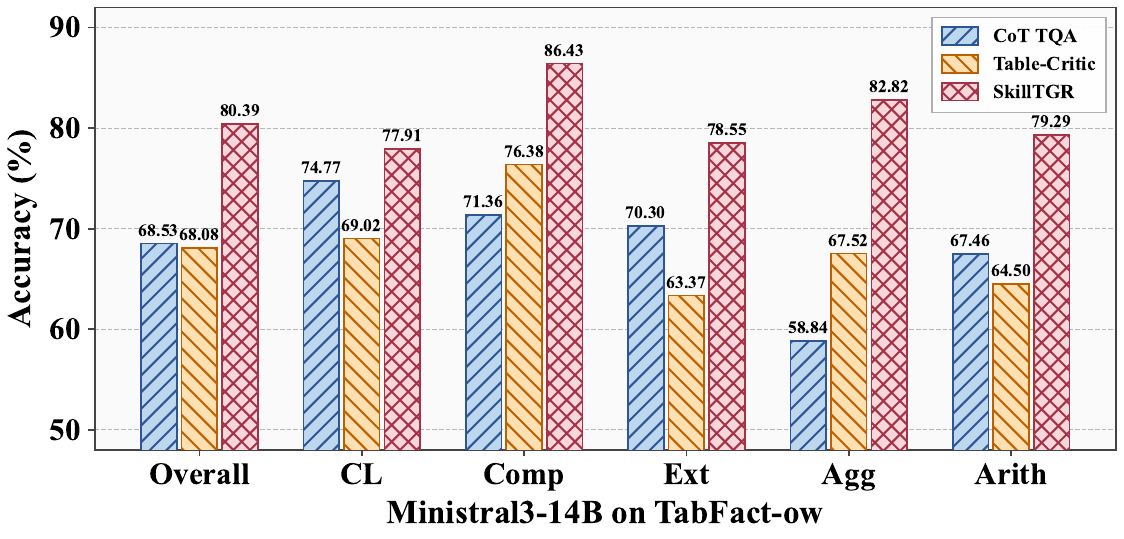}
    \label{fig:scaling_ministral}
\end{subfigure}
}
\vspace{-1.0em}
\caption{Scaling analysis on Qwen3.5-27B (left) and Ministral3-14B (right) across two benchmarks.}
\label{fig:scaling_results}
\end{figure*}

\begin{table*}[ht]
\centering
\caption{Computational efficiency comparison between Table-Critic and our SkillTGR. \textbf{Input}/\textbf{Output}/\textbf{Total} tokens and \textbf{Inference} latency (second) are averaged per question, while \textbf{Table-to-Graph} reports the one-time graph construction overhead.}
\label{tab:efficiency}
\scalebox{0.97}{
\begin{tabular}{clcccc|cccc|cc}
\toprule
\multirow{2}{*}[-0.8ex]{\textbf{Dataset}} & \multicolumn{1}{c}{\multirow{2}{*}[-0.8ex]{\textbf{Method}}} & \multicolumn{4}{c|}{\textbf{Qwen3.5-9B}}              & \multicolumn{4}{c|}{\textbf{Ministral3-8B}} & \multicolumn{2}{c}{\textbf{Table-to-Graph}}           \\ \cmidrule{3-12} 
                         & \multicolumn{1}{c}{}                        & \textbf{Input} & \textbf{Output} & \textbf{Total} & \textbf{Inference} & \textbf{Input} & \textbf{Output} & \textbf{Total} & \textbf{Inference} & \textbf{Avg (s)} & \textbf{Total (s)} \\ \midrule
\multirow{3}{*}[-0.8ex]{\textbf{WikiTQ-ow}}  
                            & Table-Critic &34,287  &1,019 &35,306  &11.78  &29,758  &894  &30,652  &6.64  &0.0 &0.0 \\
                            & SkillTGR     &26,843  &827 &27,670  &8.82  &23,848  &753  &24,601  &4.51  &0.025 &10.48  \\ \cmidrule{2-12} 
                            & \multicolumn{1}{c}{$\triangle$}   &\textcolor{teal}{$\downarrow$21.71\%}  &\textcolor{teal}{$\downarrow$18.84\%}
                            &\textcolor{teal}{$\downarrow$21.63\%}
                            &\textcolor{teal}{$\downarrow$25.13\%}  &\textcolor{teal}{$\downarrow$19.86\%}  &\textcolor{teal}{$\downarrow$15.77\%}
                            &\textcolor{teal}{$\downarrow$19.74\%}
                            &\textcolor{teal}{$\downarrow$32.08\%}  &\textcolor{red}{$\uparrow$0.025} &\textcolor{red}{$\uparrow$10.48}  \\ \midrule
\multirow{3}{*}[-0.8ex]{\textbf{TabFact-ow}} 
                            & Table-Critic &31,932  &994 &32,926  &9.01  &27,461  &873  &28,334  &4.33  &0.0 &0.0  \\
                            & SkillTGR     &25,749  &818 &26,567  &6.57  &22,495  &746  &23,241  &3.04  &0.016 &4.79  \\ \cmidrule{2-12} 
                            & \multicolumn{1}{c}{$\triangle$}  &\textcolor{teal}{$\downarrow$19.36\%}  &\textcolor{teal}{$\downarrow$17.71\%}
                            &\textcolor{teal}{$\downarrow$19.31\%}
                            &\textcolor{teal}{$\downarrow$27.08\%}  &\textcolor{teal}{$\downarrow$18.08\%}  &\textcolor{teal}{$\downarrow$14.55\%}
                            &\textcolor{teal}{$\downarrow$17.98\%}
                            &\textcolor{teal}{$\downarrow$29.79\%}  &\textcolor{red}{$\uparrow$0.016} &\textcolor{red}{$\uparrow$4.79}  \\ \bottomrule
\end{tabular}
}
\end{table*}

\textbf{Third}, although the strong baseline Table-Critic achieves competitive performance on Qwen3.5-9B, the overall accuracy drops by 18.12\% and 19.56\% on two datasets when utilizing Ministral3-8B. This observation suggests that its iterative self-refinement paradigm relies heavily on the intrinsic reasoning ability of LLMs. In contrast, SkillTGR exhibits stronger robustness across different backbones by decomposing reasoning into explicit graph traversal operations guided by skills augmentation. Such a design reduces reliance on the LLM's reasoning capability and achieves a better trade-off between performance and robustness, with a substantial margin of 11.54\% in overall accuracy and 12.16\% in operation-wise accuracy on Ministral3-8B.

\textbf{Finally}, to investigate the impact of backbone size, we conduct scaling experiments on Qwen3.5-27B and Ministral3-14B against strong baselines CoT TQA and Table-Critic. As shown in Fig.~\ref{fig:scaling_results}, although all methods benefit from increased model capacity, our SkillTGR still consistently achieves superior performance under both overall and operation-wise evaluation, yielding average improvements of 5.89\% and 5.13\%. Notably, SkillTGR attains the highest accuracies of 91.96\%, 84.49\%, 92.18\% and 85.80\% on the \emph{Comparison}, \emph{Extremum}, \emph{Aggregation} and \emph{Arithmetic} operations, which further validate that attributed graph modeling effectively facilitates cross-row relational reasoning and multi-step numerical computation, while the SkillBank enables experience transfer and pattern reuse across analogous instances.

\begin{table*}[ht]
\centering
\caption{Ablation study of SkillTGR, evaluating the contributions of Table Graph design and Skill-related components.}
\label{tab:ablation_study}
\scalebox{1.0}{
\begin{tabular}{l|cccccc|cccccc}
\toprule
\multicolumn{1}{c}{\multirow{2}{*}[-0.8ex]{\textbf{Method}}} & \multicolumn{6}{|c|}{\textbf{WikiTQ-ow}}                                                      & \multicolumn{6}{c}{\textbf{TabFact-ow}}                                     \\ \cmidrule{2-13} 
                                & \textbf{Overall} & \textbf{CL} & \textbf{Comp} & \textbf{Ext} & \textbf{Agg} & \textbf{Arith} & \textbf{Overall} & \textbf{CL} & \textbf{Comp} & \textbf{Ext} & \textbf{Agg} & \textbf{Arith} \\ \midrule
\multicolumn{1}{c|}{\textbf{SkillTGR (Full model)}}                            &\textbf{80.39}	&\textbf{82.34} &\textbf{79.84}	&\textbf{77.20} &\textbf{80.74}	&\textbf{74.46}                &\textbf{86.17}                  &\textbf{86.27}             &\textbf{88.44}               &\textbf{79.87}              &\textbf{89.29}              &\textbf{83.43}                \\ \midrule
\rowcolor{gray!20} \multicolumn{13}{l}{\textbf{\emph{Table Graph Ablation}}} \\

\quad \emph{w/o Reasoner} &76.04	&78.54	&72.61	&74.11	&76.11	&69.26 &80.09	&81.44	&78.89	&73.60	&82.65	&78.11 \\ 

\quad \emph{w/o Planner\&Executor} &71.48	&74.69	&70.28	&68.10	&70.32	&67.97 &74.36	&77.65	&72.86	&68.65	&73.81	&73.37 \\

\quad \emph{w/o Table-to-Graph} &66.78	&73.88	&63.82	&63.02	&62.08	&62.77 &68.68	&73.46	&68.34	&65.68	&64.97	&65.68 \\ \midrule

\rowcolor{gray!20} \multicolumn{13}{l}{\textbf{\emph{Skill-related Ablation}}} \\ 

\quad \emph{w/o Hierarchy} &79.10	&81.22	&78.29	&76.58	&79.17	&72.29 &85.03	&85.49	&86.93	&78.55	&87.93	&82.25 \\ 

\quad \emph{w/o Skill Evolving} &78.91	&81.28	&78.04	&75.65	&79.10	&71.86 &84.93	&85.23	&87.44	&78.88	&87.76	&81.66 \\

\quad \emph{w/o Planner Aug.} &77.95	&80.60	&76.49	&75.04	&77.74	&71.43 &82.71	&83.27	&82.91	&76.9	&85.54	&80.47 \\

\quad \emph{w/o Executor Aug.} &77.83	&80.47	&76.23	&74.88	&77.54	&72.29 &82.36	&83.53	&81.41	&76.24	&84.69	&81.07 \\ 

\quad \emph{w/o Reasoner Aug.} &77.76	&80.53	&76.74	&75.19	&77.26	&70.56 &82.11	&82.88	&80.90	&75.91	&84.86	&81.66 \\ 

\quad \emph{w/o SkillBank} &76.59	&78.98	&73.39	&74.58	&76.79	&69.70 &80.58	&81.57	&79.90	&73.93	&83.16	&79.88\\ \bottomrule
\end{tabular}
}
\end{table*}

\begin{table*}[ht]
\centering
\caption{Performance comparison on short and long tables under row-based and token-based partitioning.}
\label{tab:long_tables}
\scalebox{0.98}{
\begin{tabular}{llcccccc|cccccc}
\toprule
\multicolumn{1}{c}{\multirow{2}{*}[-0.8ex]{\textbf{Partition}}} & \multicolumn{1}{c}{\multirow{2}{*}[-0.8ex]{\textbf{Method}}} & \multicolumn{6}{c|}{\textbf{Qwen3.5-9B}} & \multicolumn{6}{c}{\textbf{Ministral3-8B}}                                     \\ \cmidrule{3-14} 
                                & \multicolumn{1}{c}{} & \textbf{Overall} & \textbf{CL} & \textbf{Comp} & \textbf{Ext} & \textbf{Agg} & \textbf{Arith} & \textbf{Overall} & \textbf{CL} & \textbf{Comp} & \textbf{Ext} & \textbf{Agg} & \textbf{Arith} \\ \midrule
\multirow{2}{*}[-0.3ex]{\textbf{Row$<$50 (4,003)}}
&Table-Critic                    &80.11 &\textbf{83.66} &78.03 &76.80 &79.30 &73.56          &62.30 &68.07 &60.40 &62.25 &57.78 &54.33          \\
&SkillTGR                       &\textbf{80.71} &82.71 &\textbf{80.35} &\textbf{77.29} &\textbf{81.13} &\textbf{74.52}          &\textbf{73.27} &\textbf{76.07} &\textbf{76.01} &\textbf{72.06} &\textbf{71.07} &\textbf{66.83}          \\ \midrule

\multirow{2}{*}[-0.3ex]{\textbf{Row$\geq$50 (341)}}
&Table-Critic                    &71.26 &75.19 &70.73 &64.86 &69.16 &69.57          &49.56 &56.39 &53.66 &56.76 &37.38 &47.83          \\
&SkillTGR                       &\textbf{76.54} &\textbf{78.20} &\textbf{75.61} &\textbf{75.68} &\textbf{75.70} &\textbf{73.91}          &\textbf{69.50} &\textbf{72.18} &\textbf{68.29} &\textbf{70.27} &\textbf{67.29} &\textbf{65.22}          \\ \midrule \midrule
                                
\multirow{2}{*}[-0.3ex]{\textbf{Token$<$4k (4202)}}                                
&Table-Critic                    &79.70 &\textbf{83.14} &78.11 &76.42 &78.81 &73.57          &61.90 &67.89 &60.00 &62.18 &57.05 &53.74          \\
&SkillTGR                       &\textbf{80.41} &82.36 &\textbf{79.73} &\textbf{77.22} &\textbf{80.84} &\textbf{74.45}          &\textbf{73.11} &\textbf{75.90} &\textbf{75.41} &\textbf{72.15} &\textbf{70.95} &\textbf{66.52}          \\ \midrule

\multirow{2}{*}[-0.3ex]{\textbf{Token$\geq$4k (142)}}
&Table-Critic                    &71.13 &78.33 &58.82 &64.71 &70.45 &50.00          &43.66 &46.67 &52.94 &52.94 &31.82 &50.00          \\
&SkillTGR                       &\textbf{79.58} &\textbf{81.67} &\textbf{82.35} &\textbf{76.47} &\textbf{77.27} &\textbf{75.00}          &\textbf{69.01} &\textbf{71.67} &\textbf{70.59} &\textbf{64.71} &\textbf{65.91} &\textbf{75.00}          \\ \bottomrule
\end{tabular}
}
\end{table*}

\subsection{Efficiency Analysis}
To further assess computational efficiency, we compare SkillTGR with the strong baseline Table-Critic in terms of token consumption and inference latency. As reported in Table \ref{tab:efficiency}, our SkillTGR consistently outperforms Table-Critic across both datasets and LLM backbones, reducing token consumption and inference latency by 19.76\% and 27.64\% on average, respectively. The underlying reason is that Table-Critic relies on iterative critique-refinement interactions until converging to correct solutions, which repeatedly propagates the full linearized table as context and generates redundant intermediate responses, thus resulting in higher computational overhead. In contrast, our SkillTGR confines reasoning to local evidence subgraphs through dynamic graph traversal, while leveraging historical reasoning experience to augment inference with reusable skills rather than reasoning from scratch, thereby improving efficiency substantially. Although SkillTGR requires an additional table-to-graph construction, the average overhead is merely 0.025 and 0.016 seconds per table on two datasets. As a one-time offline preprocessing step, it is negligible compared with overall inference costs. Consequently, our SkillTGR achieves a more favorable accuracy-efficiency trade-off while delivering superior reasoning performance.

\subsection{Ablation Study}
To dissect component contributions, we conduct an ablation study using Qwen3.5-9B with two groups of ablations on both datasets, as reported in Table~\ref{tab:ablation_study}.

For the \textbf{Table Graph Ablation}, \emph{w/o Reasoner} removes explicit reasoning path generation with overall and operation-wise decline of 5.22\% and 5.72\%, verifying graph traversal reasoning enhances evidence aggregation for complex multi-step inference. Moreover, \emph{w/o Planner\&Executor} eliminates dynamic chain planning and evidence subgraph retrieval with substantial degradation of 11.06\% on cross-row operations, indicating dynamic evidence retrieval is crucial for filtering noise and pinpointing supporting evidence. \emph{w/o Table-to-Graph} further discards both table graph construction and dynamic graph traversal reasoning, retaining only skill-augmented reasoning. This variant exhibits the largest degradation of 15.55\% and 15.88\% on overall and operation-wise accuracy, confirming that attributed table graphs effectively preserve structural semantics and mitigate the ``lost-in-the-middle'' issue.

For the \textbf{Skill-related Ablation}, \emph{w/o Hierarchy} causes decreases of 1.38\% operation-wise performance when removing the operation-wise partitioning and success/failure bifurcation, verifying that the hierarchical organization reduces cross-skill interference and improves retrieval precision. In addition, \emph{w/o Skill Evolving} disables continual merging for skill refinement with 1.36\% overall and 1.56\% operation-wise drops, suggesting skill evolution benefits maintaining high-quality experience accumulation. Furthermore, \emph{w/o Planner Aug.}, \emph{w/o Executor Aug.} and \emph{w/o Reasoner Aug.} remove contrastive skill augmentation for the Planner, Executor and Reasoner, which consistently incur significant decline with 2.95\%--3.35\% overall and 3.21\%--3.60\% operation-wise performance, demonstrating that contrastive augmentation from successful and failed skills facilitates reasoning with positive guidance and negative constraint. Finally, \emph{w/o SkillBank} completely removes the Skill-Augmented Self-Evolution module, yielding the largest degradation of 4.70\% and 5.06\% on overall and operation-wise accuracy, which confirms the continual self-evolution mechanism effectively accumulates transferable experience and enhances reasoning capability beyond the conventional reason-from-scratch paradigm.

In summary, each module of our SkillTGR contributes to the entire performance on the operation-wise table reasoning.

\begin{figure*}[ht]
    \centering
    \includegraphics[width=1.0\linewidth]{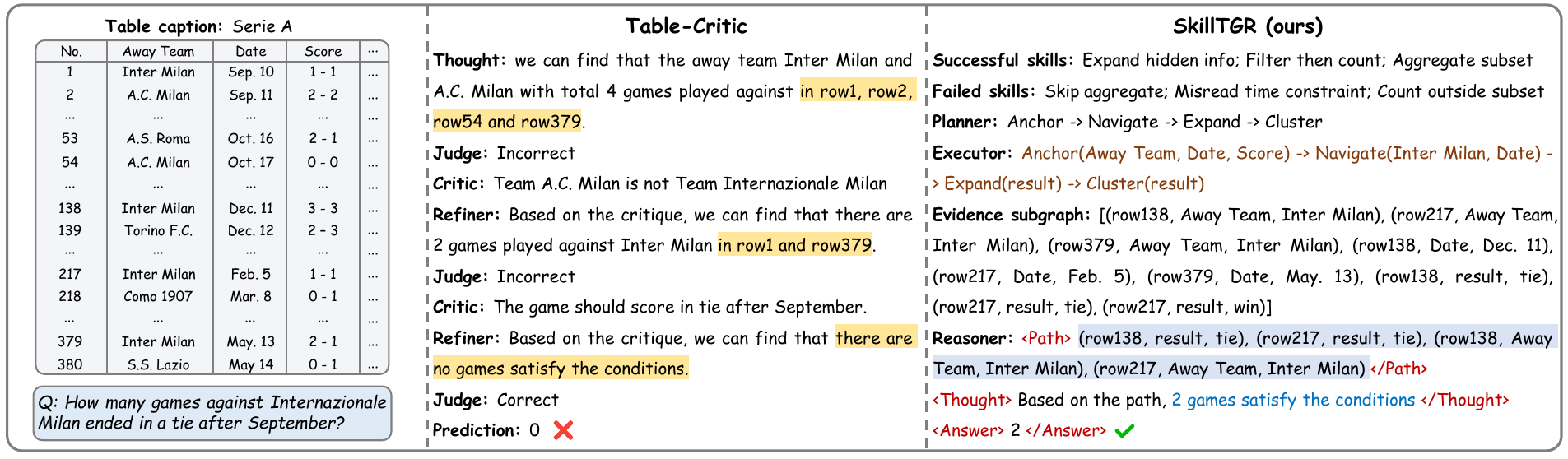}
    \caption{Case study.}
    \label{fig:case_study}
\end{figure*}

\begin{figure*}[ht]
\centering

\begin{subfigure}[t]{0.32\linewidth}
    \centering
    \includegraphics[width=\linewidth]{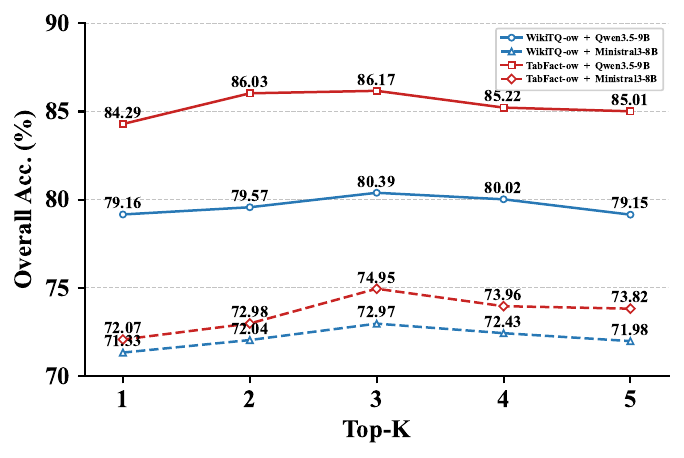}
    \label{fig:topk}
\end{subfigure}
\hfill
\begin{subfigure}[t]{0.32\linewidth}
    \centering
    \includegraphics[width=\linewidth]{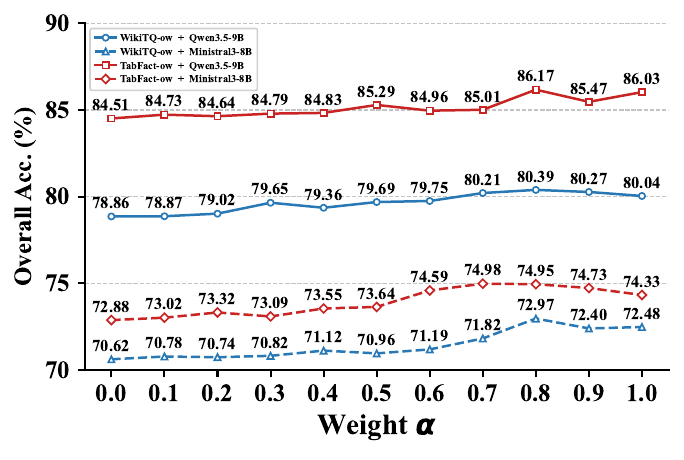}
    \label{fig:alpha}
\end{subfigure}
\hfill
\begin{subfigure}[t]{0.32\linewidth}
    \centering
    \includegraphics[width=\linewidth]{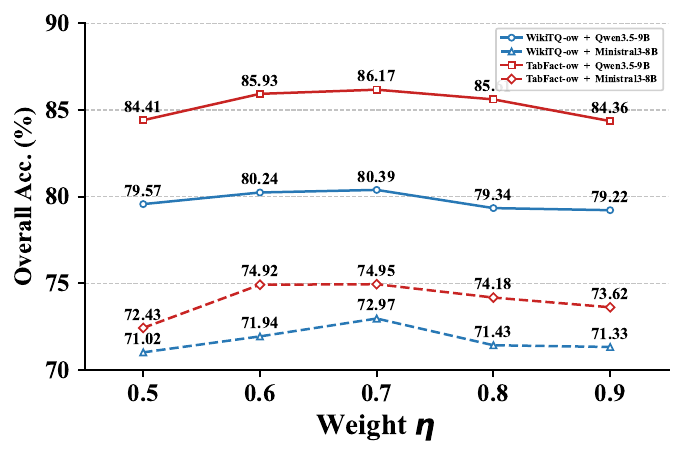}
    \label{fig:eta}
\end{subfigure}
\vspace{-1.0em}
\caption{Hyperparameter sensitivity analysis}
\label{fig:hyperparameter}
\end{figure*}

\subsection{Table Size Analysis}
To evaluate robustness across different table scales, we partition WikiTQ-ow tables (since all TabFact-ow tables are short) into short and long subsets along two dimensions: row-based (Row$\geq$50 as long) and token-based (Token$\geq$4k as long), yielding 341 and 142 long-table questions respectively.

As shown in Table~\ref{tab:long_tables}, both methods achieve competitive performance on short tables under Qwen3.5-9B, whereas the advantage of SkillTGR becomes increasingly pronounced on long tables. Under row-based partitioning, our SkillTGR outperforms Table-Critic by 5.28\% and 19.94\% in overall accuracy on Qwen3.5-9B and Ministral3-8B respectively, with gains further increasing to 8.45\% and 25.35\% under token-based partitioning. In particular, the improvements are concentrated on \emph{Comparison}, \emph{Extremum}, \emph{Aggregation} and \emph{Arithmetic} questions, which require extensive cross-row evidence integration and multi-step numerical reasoning. SkillTGR achieves average operation-wise gains of 12.75\% and 19.45\% on these complex questions under row-based and token-based partitioning respectively. Such disparities stem from the fact that Table-Critic still relies on linearized tables and reason-from-scratch inference, making it vulnerable to inefficient evidence retrieval on large tables. In contrast, SkillTGR preserves explicit table structures through attributed graphs and further leverages reusable skills for experience-augmented reasoning, leading to stronger scalability for complex reasoning over long tables.

\subsection{Case Study}
To intuitively illustrate the superiority of our SkillTGR in complex structured reasoning, we present an aggregation case study against the baseline Table-Critic, as shown in Fig.~\ref{fig:case_study}.

Specifically, this question requires counting games that simultaneously satisfy three constraints: the opponent is Internazionale Milan, the match occurs after September, and the final result is a tie. Since the target evidence rows are deeply buried within lengthy and noisy contexts, the task poses significant challenges for long-context retrieval and multi-step aggregation. Although Table-Critic progressively corrects errors through iterative critique-refinement, its linearized table representation is inherently susceptible to the ``lost-in-the-middle'' issue, causing crucial evidence overlooked and leading to an incorrect answer. In contrast, SkillTGR transforms tables into attributed graphs and performs dynamic graph traversal reasoning to efficiently retrieve evidence subgraphs while filtering irrelevant information. Furthermore, contrastive augmentation with successful and failed skills equips the multi-agent framework with transferable reasoning experience, enabling more accurate retrieval planning and aggregation reasoning, thereby producing the correct prediction.

\subsection{Hyperparameter sensitivity analysis}
To investigate the impact of the key hyperparameters in the hybrid skill retrieval, we conduct a sensitivity study on both datasets across different LLMs, as illustrated in Fig.~\ref{fig:hyperparameter}.

Specifically, the Top-k determines the number of successful and failed skills injected for contrastive augmentation. Smaller values provide insufficient experience coverage, whereas larger values introduce irrelevant noise. SkillTGR achieves the best performance at k=3, indicating an optimal balance between experience diversity and retrieval noise. The weight coefficient $\alpha$ balances dense semantic similarity and BM25-based lexical matching. A lower $\alpha$ favors lexical alignment but overlooks semantically similar skills, while a higher $\alpha$ improves semantic generalization at the expense of discriminative ability. The optimal $\alpha=0.8$ suggests that semantic-dominant retrieval with moderate lexical constraints best captures transferable skill patterns. Finally, the threshold $\eta$ filters retrieved candidates by relevance. A lower threshold introduces noisy skills, whereas a higher threshold over-filters useful ones and reduces coverage. The best performance is achieved at $\eta=0.7$, demonstrating the importance of balancing skill quality and coverage.

\section{Conclusion}
In this paper, we introduce a novel \emph{Operation-wise TableQA} task together with two refined benchmarks, to investigate how LLMs perform across a fine-grained question taxonomy with five core operations rather than the limited overall assessment. Based on this, we propose the SkillTGR framework to represent tables as attributed graphs for dynamic graph traversal reasoning, mitigating both structural semantic loss and the ``lost-in-the-middle'' issue inherent in conventional table linearization approaches. Moreover, our SkillTGR constructs a hierarchical SkillBank to distill historical trajectories into reusable skills for contrastive augmented reasoning, facilitating the continual self-evolution across diverse operation-wise instances. Extensive experiments demonstrate that our method significantly outperforms competitive baselines in both overall and operation-wise evaluations, while substantially reducing token consumption and inference latency, validating both the effectiveness and efficiency of our proposed framework.

\section*{Acknowledgment}
In accordance with IEEE guidelines, we disclose the usage of AI tools in this work. GPT-5 was employed in the construction of the WikiTQ-ow and TabFact-ow datasets (described in Section 4, Experimental Settings) to assist with question categorization, operation-type annotation, and QA pair generation. All AI-generated annotations and data were subsequently reviewed and verified by domain experts to ensure correctness and reliability. In addition, LLMs are used for grammar checking and linguistic polishing of the manuscript.

\balance
\bibliographystyle{IEEEtran}
\bibliography{reference}

\end{document}